%% file: main.tex
\documentclass[acmtog,nonacm,screen,authorversion]{acmart}

\AtBeginDocument{%
  }

\setcopyright{acmlicensed}
\copyrightyear{2018}
\acmYear{2018}
\acmDOI{XXXXXXX.XXXXXXX}
\acmConference[Conference acronym 'XX]{Make sure to enter the correct
  conference title from your rights confirmation email}{June 03--05,
  2018}{Woodstock, NY}
\acmISBN{978-1-4503-XXXX-X/2018/06}

\usepackage{booktabs}
\usepackage{multirow}
\usepackage{array}
\usepackage{graphicx}
\usepackage{subcaption}
\usepackage{amsmath}
\usepackage{float}
\usepackage{algorithm}
\usepackage{algorithmic}
\usepackage{xcolor}
\usepackage{pifont}
\usepackage{enumitem}

\newcommand{\cmark}{\textcolor{green!60!black}{\ding{51}}}
\newcommand{\xmark}{\textcolor{red!70!black}{\ding{55}}}

\makeatletter
\newcommand{\projectpage}[1]{%
  \gdef\@projectpage{\href{#1}{\nolinkurl{#1}}}%
}
\def\@projectpage{}
\newcommand{\@mkprojectpage}{%
  \ifx\@projectpage\@empty\else
    \global\setbox\mktitle@bx=\vbox{%
      \noindent\unvbox\mktitle@bx\par
      \medskip
      \noindent\small\textbf{Project Page:}\enspace\@projectpage\par
      \medskip
    }%
  \fi
}
\let\Lumera@orig@mkauthors\@mkauthors
\renewcommand{\@mkauthors}{%
  \Lumera@orig@mkauthors
  \@mkprojectpage
}
\renewcommand\@ACM@checkaffil{}
\makeatother
\acmSubmissionID{1141}

\title{Engine-Native Editable 3D World Reconstruction with Objects and Lighting}

\author{Junhao Chen}
\authornote{Equal Contribution.}
\affiliation{%
  \institution{Tsinghua University}
}
\email{yisuanwang@gmail.com}

\author{Xinghao Chen}
\authornotemark[1]
\affiliation{%
  \institution{Nankai University}
}
\email{Xinghao-Chen@outlook.com}

\author{Henghaofan Zhang}
\authornotemark[1]
\affiliation{%
  \institution{University of Electronic Science and Technology of China}
}
\email{hhfzhang@outlook.com}

\author{Zihao Qiao}
\affiliation{%
  \institution{Sun Yat-sen University}
}
\email{Qiao20050212@gmail.com}

\author{Saining Zhang}
\affiliation{%
  \institution{Nanyang Technological University}
}
\email{saining002@e.ntu.edu.sg}

\author{Yongzhi Li}
\affiliation{%
  \institution{Nanyang Technological University}
}
\email{YONGZHI001@e.ntu.edu.sg}

\author{Ruqi Huang}
\affiliation{%
  \institution{Tsinghua University}
}
\email{ruqihuang@sz.tsinghua.edu.cn}

\author{Sisi Li}
\affiliation{%
  \institution{Ophilus.AI}
}
\email{sisi@ophilus.ai}

\author{Yimin Sheng}
\affiliation{%
  \institution{Ophilus.AI}
}
\email{yimin@ophilus.ai}

\author{Jianyi Zhu}
\affiliation{%
  \institution{Ophilus.AI}
}
\email{chris@ophilus.ai}

\author{Hao Zhao}
\authornote{Corresponding Author.}
\affiliation{%
  \institution{Tsinghua University}
}
\email{zhaohao@air.tsinghua.edu.cn}

\renewcommand{\shortauthors}{Chen et al.}
\projectpage{https://haidilao0328.github.io/Lumera/}

\ccsdesc[500]{Computing methodologies~Computer graphics}
\ccsdesc[300]{Computing methodologies~Scene understanding}
\ccsdesc[300]{Computing methodologies~Spatial reasoning}

\keywords{Editable 3D scenes, game-engine data, parametric lighting, scene parsing, single-image 3D reconstruction}

\begin{document}
\input{sections/abstract}

\begin{teaserfigure}
  \centering
  \includegraphics[width=\textwidth]{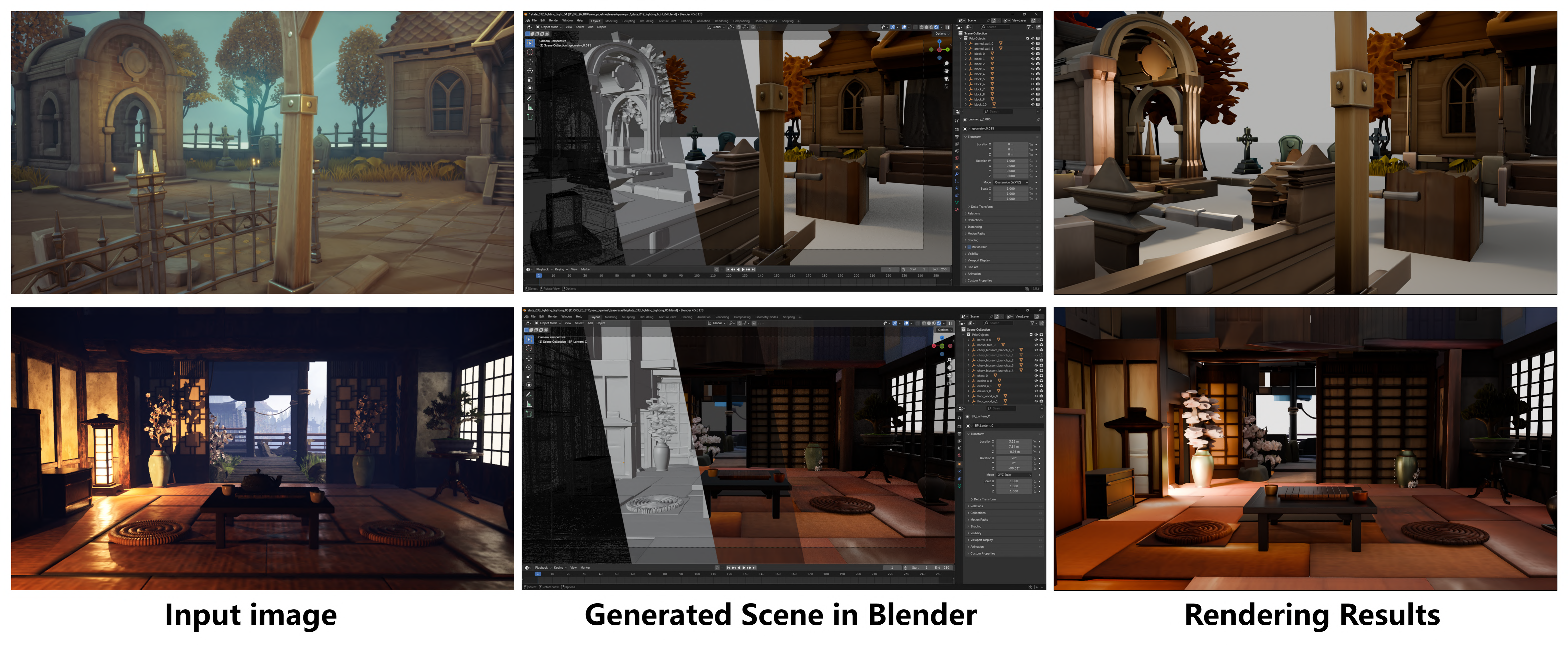}
  \caption{Given a single image, Lumera parses object-level
  3D boxes and engine-native parametric lights, reconstructs per-object meshes,
  estimates an HDR environment probe, and assembles an editable scene that can be
  loaded in Blender or UE5.}
  \Description{Teaser figure showing the proposed pipeline for object boxes,
  parametric lights, per-object meshes, and editable scene assembly from a
  single open-world game-scene image.}
\end{teaserfigure}

\maketitle

\input{sections/intro}
\input{sections/related}
\input{sections/dataset}
\input{sections/method}
\input{sections/experiments}
\input{sections/conclusion}

\clearpage

\bibliographystyle{ACM-Reference-Format}
\bibliography{main}

\clearpage
\input{sections/postref_figures}

\clearpage
\appendix
\input{sections/appendix}

\end{document}

%% file: sections/abstract.tex
\begin{abstract}
Editable 3D scene creation requires object instances and lights that can be inspected, moved, and imported into standard engines, yet existing single-image methods largely stop at room-scale geometry, baked/global illumination, or text-driven generation. We introduce \textbf{Lumera} (\emph{Light-aware Unified Engine-native Reconstruction and Assembly}), a benchmark and reference pipeline for engine-native, light-aware 3D scene parsing from a single image. \textbf{Lumera-2K} is built from $2{,}513$ UE5 projects and provides $3.73$M components, $63$M object instances, $102.6$K engine-native parametric lights, and $95.1$K camera views. On this data, \textbf{Lumera-Box} and \textbf{Lumera-Light} adapt VLM to parse object boxes and parametric light tuples $(x,y,z,r,g,b,I)$, which are assembled with per-object mesh reconstruction, HDR environment estimation, and a bounded agentic refinement loop. In a sanitized box benchmark against DetAny3D, SpatialLM, N3D-VLM, and WildDet3D, Lumera-Box obtains the strongest overall detection, geometry, semantic, and layout scores (merged mAP$\uparrow$ $0.1141$, IoU-B$\uparrow$ $0.2472$, F-score$\uparrow$ $0.2762$), while WildDet3D remains stronger on anchor recall$\uparrow$. For lights, Lumera-Light recovers almost all non-empty scenes (recall$\uparrow$ $0.998$) but remains limited at individual-light localization (F1$\uparrow$ $0.209$ at $0.5$\,m); matched lights have median position error$\downarrow$ $0.261$\,m, median $\Delta E_{2000}\downarrow$ $4.59$, and intensity Pearson $r\uparrow=0.628$. These results establish parametric lights as a measurable editable-scene target and expose remaining bottlenecks in relation structure, light recall/intensity, and cross-engine generalization.
\end{abstract}

%% file: sections/intro.tex
\section{Introduction}
\label{sec:intro}

Turning a single image into an editable 3D scene is a long-standing graphics
goal with direct value for game production, virtual production, simulation,
robotics, and asset creation. More recently, it has also become a prerequisite
for \emph{controllable video generation}~\cite{ma2025controllable,chen2026dancetogether,xue2025human}:
a growing class of video diffusion
models~\cite{das,kim2025videofrom3d,Huang_2026_CVPR_CineScene} now condition on
an explicit 3D scene, including coarse geometry, object placements, camera
trajectories, and 3D tracking videos, to achieve precise camera control,
object manipulation, and temporal consistency. Whether the downstream tool is
Blender, UE5, or a 3D-conditioned video model, the requirement is the same: the
output must be more than plausible pixels or a fused geometry proxy. Users need
object instances that can be moved, meshes that can be replaced or refined, and
lights whose positions, colors, and intensities can be inspected and edited.
This becomes especially challenging in game-scale scenes, where a single view
may contain dense foreground assets, mixed indoor/outdoor structure, large
metric extent, complex occlusion, and many local lights interacting with global
illumination.

Existing methods cover important parts of this problem but leave the editable
scene representation incomplete. Multi-instance image-to-3D
systems~\cite{huang2025midi,yao2025cast,zhai2024echoscene, chen2026videoworldturningmonocular} model object-centric
scenes, but they are mainly demonstrated on room-scale settings and do not
predict engine-native lights. Feed-forward geometry
models~\cite{yang2025depthanything3,wang2025vggt,leroy2024mast3r} produce
strong depth, camera, and point-cloud priors, yet their outputs are not
editable object and light entities. Agentic and text-driven
generators~\cite{xia2026sage,pfaff2026scenesmith,yin2026viga,zhang2026code2worlds}
can create or edit scenes through tool calls, but they generally synthesize a
new scene from a prompt rather than parse the depicted one. Lighting-estimation
methods recover environment maps, spherical harmonics, or intrinsic components,
which are useful for relighting but cannot be imported as independent point,
spot, rect, or directional light components. Symmetrically, 3D-conditioned
video generators~\cite{das,kim2025videofrom3d,Chen_2026_CVPR_hvg3d} assume that a suitable 3D scene
condition is already available, yet producing such a condition for a depicted
game-scale image with many instances, mixed indoor/outdoor extent, and
engine-native lights is precisely what current single-image parsers cannot do.

We argue that the missing step is to treat image-to-scene reconstruction as
\emph{game-engine structured parsing}. Game engines already store the entities
that downstream tools need: object components, transforms, cameras, materials,
SkyLight probes, and countable parametric lights. We therefore introduce
\textbf{Lumera} (\emph{Light-aware Unified Engine-native Reconstruction and
Assembly}), a benchmark and reference pipeline for engine-native, light-aware 3D
scene parsing from a single image and its recovered point cloud. Lumera uses
parsing as the front end of engine-native reconstruction: it predicts object
boxes, text labels, and parametric lights as structured entities, reconstructs
foreground objects and an HDR environment, and assembles them into an editable
engine scene. The core idea is to factor the problem around editable entities:
predict boxes and lights as structured tokens, reconstruct each foreground
object independently, assemble the scene in an engine representation, and use an
agent only as a bounded editor over allowed fields.

The Lumera pipeline has four coupled parts, each tied to this entity-centric
formulation. First, \textbf{Lumera-2K} provides the supervision that existing
datasets lack: $2{,}513$ UE5 projects with $3.73$M components, $63$M object
instances, $102.6$K parametric lights, and $95.1$K camera views. Second,
\textbf{Lumera-Box} adapts SpatialLM~\cite{wang2025spatiallm} to game-domain
object boxes, producing oriented 3D boxes and labels from a colored point cloud.
Third, \textbf{Lumera-Light} uses a separate SpatialLM checkpoint to emit
engine-light tuples $(x,y,z,r,g,b,I)$, making local illumination a measurable
scene entity rather than a baked appearance effect. Fourth, the assembly stage
projects boxes to image masks, reconstructs per-object meshes with
SAM3D~\cite{vorobyev2026sam3d}, estimates an HDR environment with
IntrinsicHDR~\cite{dille2024intrinsichdr}, and runs a VIGA-inspired
Generator/Verifier loop~\cite{yin2026viga} whose geometry stage may edit only
object yaw/scale and whose lighting stage may edit only existing lights,
environment strength, and exposure.

Experiments show that this formulation is already useful and also exposes clear
remaining bottlenecks. On the merged Lumera-2K val+test split, Lumera-Box
outperforms DetAny3D, zero-shot SpatialLM, N3D-VLM, and WildDet3D on the main
detection and geometry metrics, including mAP$\uparrow$ $0.1141$,
IoU-B$\uparrow$ $0.2472$, and F-score$\uparrow$ $0.2762$, while also improving
scene semantics and graph consistency (scene-semantic score$\uparrow$ $0.3827$,
GCC$\uparrow$ $0.6676$). WildDet3D remains stronger on SRF$\uparrow$ and
anchor recall$\uparrow$, indicating that relation recovery is not solved. For
lights, Lumera-Light reaches nonempty-scene recall$\uparrow$ $0.998$, but
individual-light localization is still difficult: F1$\uparrow$ is $0.209$ at a
$0.5$\,m threshold, median position error$\downarrow$ is $0.261$\,m, median
$\Delta E_{2000}\downarrow$ is $4.59$, and intensity Pearson
$r\uparrow$ is $0.628$. A bounded refinement case study improves a
55-instance scene, but fails to rescue poor initial outdoor parses, confirming
that the agentic loop should be evaluated as a constrained editor rather than as
a substitute for structured parsing.

Our contributions are:
\begin{itemize}[leftmargin=0pt, itemsep=0pt, topsep=0pt]
\item \textbf{A new task and dataset.}
We define game-scale, light-aware image-to-scene parsing, where the target
output includes object boxes, object meshes, engine-native parametric lights,
and an HDR environment probe. We introduce \textbf{Lumera-2K}, a
UE5-derived dataset with dense object, camera, HDRI, and countable light
annotations for editable-scene supervision.

\item \textbf{A light-aware parsing, reconstruction, and assembly pipeline.}
We propose the \textbf{Lumera} pipeline, which combines two SpatialLM-based parsers,
\textbf{Lumera-Box} and \textbf{Lumera-Light}, with box-guided per-object
mesh reconstruction, HDR environment estimation, and stage-aware agentic
refinement with executable scope validation and rollback.

\item \textbf{A benchmark for game-scale editable scenes.}
We report aligned 3D box benchmarks against DetAny3D, SpatialLM, N3D-VLM, and
WildDet3D, introduce quantitative metrics for parametric light prediction, and
provide post-reference qualitative audits for point-cloud parsing and
reconstruction comparisons.
\end{itemize}

%% file: sections/related.tex
\input{sections/table_dataset_complexity}

\section{Related Work}
\label{sec:related}

\subsection{Scene Reconstruction and Generation}
\label{sec:related_recon}
Neural radiance fields and 3D Gaussian
splatting~\cite{mildenhall2020nerf,mueller2022instant,kerbl2023gaussiansplatting,huang2024gs2d,lu2024scaffoldgs,liu2024rip}
have made view synthesis highly effective, but their outputs are radiance
fields or pixel primitives rather than object and light entities that can be
edited in a game engine. Feed-forward 3D foundation
models~\cite{yang2024depthanythingv2,yang2025depthanything3,weng2026feedforward3deditinglearns,bochkovskii2024depthpro,yin2023metric3d,wang2024dust3r,leroy2024mast3r,chen2026ultraman,wang2025vggt,fu2024colmapfree,zhou2025ea3d,miao2026framessequencestemporallyconsistent}
provide strong depth and camera priors, multi-instance image-to-3D
methods~\cite{huang2025midi,yao2025cast,zhai2024echoscene,chen2026videoworldturningmonocular} recover
object-centric scenes, and holistic layout
methods~\cite{zhao2017physics,nie2020total3d,li2021odam,ardelean20253dsr,stekovic2023cadestate,zhang2024monst3r}
estimate room structure or global geometry, but these systems remain
primarily room-scale and do not expose engine-native light entities.
Infinite-scene and image-driven scene generation
methods~\cite{yu2024wonderjourney,chung2024luciddreamer,yu2025wonderworld,cohen2023setscene,chen2025physgen3d}
extend a plausible world from a prompt or an anchor image, but the output is
a freely synthesized scene rather than the depicted one. In contrast, our
task is to \emph{parse} an observed game-scale image into editable object and
light entities; we therefore reuse feed-forward depth and per-object mesh
reconstruction only as a geometry front end, and treat the final scene as an
explicit engine representation rather than a radiance field or a generated
world.
\subsection{Structured Data Generation}
\label{sec:related_understand}
A growing body of work generates 3D content as \emph{structured tokens}
rather than as voxels, pixels, or implicit fields, since structured outputs
are compact, interpretable, and directly editable in downstream tools.
Autoregressive mesh
generators~\cite{wang2024llama,siddiqui2024meshgpt,tang2024edgerunner,weng2026garmentgpt}
emit faces and vertices as tokens, autoregressive
rigging~\cite{sun2025drive,song2025magicarticulate,zhang2025unirig,liu2025riganything,song2026puppeteer,Sun_2026_CVPR_Animator}
predicts joints and skeleton trees as sequences, parametric
CAD/B-rep~\cite{wu2021deepcad,li2025brepgpt} and
LEGO~\cite{pun2025generating} generators emit constrained command streams,
and vector-graphics
models~\cite{yang2026omnisvg,Chen_2026_CVPR_LottieGPT} produce SVG/Lottie
code through code synthesis. The same idea has been transferred to 3D scene
understanding, where spatial-language
decoders~\cite{wang2025spatiallm,zhan2024scenescript,wang2024embodiedscan,dou2025video3dllm,xu2023pointllm}
serialize layouts and oriented boxes as tokens, and structured-caption
models~\cite{wang2025n3d,huang2026wilddet3d,wang2025spatiallm} emit
text-level scene descriptions from video or 3D inputs. In parallel,
inverse-rendering and lighting-estimation
methods~\cite{dille2024intrinsichdr} recover environment maps, spherical
harmonics, or intrinsic components, but these are continuous appearance
signals rather than countable light components. Our 3D box and parametric
light outputs are instances of this same structured-token paradigm, with
parametric lights introduced as a new structured-prediction target alongside
the more familiar mesh, rig, CAD, and SVG cases.

\subsection{Agentic 3D AIGC}
\label{sec:related_agentic}
Closed-loop and agentic 3D systems use LLM or VLM tool calls to either
generate or edit 3D scenes, typically with retrieval from large asset
libraries~\cite{deitke2023objaverse,zhang2025texverse,fu2021future}.
Prompt-driven agentic scene
generators~\cite{yang2024holodeck,xia2026sage,pfaff2026scenesmith,chi2026mansion}
synthesize a new scene from a language prompt or a high-level plan, while
agentic scene
editors~\cite{wei2024editable,huang2024blenderalchemy,hu2024scenecraft,sun2025layoutvlm,yin2026viga,zhang2026code2worlds,chen2024idea23d}
use tool calls to modify an existing scene through Blender code or layout
adjustments. Our pipeline reuses the agent idea only after structured
parsing, as a bounded editor over a whitelist of object and light fields
rather than as a free generator.

%% file: sections/table_dataset_complexity.tex
\begin{table*}[htbp]
  \centering\scriptsize
  \caption{Scale and annotation coverage of major 3D scene datasets.
  Values are reported as \textbf{total (mean per scene)};
  $\dagger$~marks numbers we derive from the public release.
  ``Lights'' is checked only when the dataset publishes
  per-light parametric annotation
  (position, type, intensity, temperature) with countable instances;
  baked HDR textures and intrinsic decompositions do not count.
  ``Labels'' is checked only when free-form / open-vocabulary text
  descriptions are provided per object.}
  \label{tab:dataset_complexity}
  \setlength{\tabcolsep}{3pt}
  \resizebox{\textwidth}{!}{%
  \begin{tabular}{@{}l l c r l l l l l c@{}}
    \toprule
    Category & Dataset & In/Outdoor & Scenes & Components total (mean) & Objects total (mean) & Lights total (mean) & Views total (mean) & HDRI total & Labels \\
    \midrule
    \multirow{5}{*}{Scan / Recon}
      & ScanNet~\cite{dai2017scannet}              & Indoor  & 1{,}513 & — & 36{,}213 (24.1) & \xmark & \cmark\,2.49M (1{,}648$\dagger$) & \xmark & \xmark \\
      & ScanNet++~\cite{yeshwanth2023scannetpp}    & Indoor  & 460     & — & —               & \xmark & \cmark\,280K DSLR (609$\dagger$) & \xmark & \cmark \\
      & Matterport3D~\cite{chang2017matterport3d}  & Indoor  & 90      & — & 50{,}811 (564.6$\dagger$) & \xmark & \cmark\,10{,}800 panorama (120) & \xmark & \cmark \\
      & Replica~\cite{straub2019replica}           & Indoor  & 18      & — & — & \xmark & \xmark & \xmark & \xmark \\
      & HM3D / HM3DSem~\cite{ramakrishnan2021hm3d} & Indoor  & 1{,}000 & — & 142{,}646 (661) & \xmark & \xmark & \xmark & \xmark \\
    \midrule
    \multirow{5}{*}{Synthetic / Designed}
      & 3D-FRONT~\cite{fu2021front}                & Indoor  & 18{,}968 & — & — (6.5$\dagger$) & \xmark & \xmark & \xmark & \xmark \\
      & Structured3D~\cite{zheng2020structured3d}  & Indoor  & 21{,}835 & — & — & \xmark & \cmark\,196{,}515 (9$\dagger$) & \xmark & \xmark \\
      & Hypersim~\cite{roberts2021hypersim}        & Indoor  & 461     & — & 58{,}693$\dagger$ (127.3) & \xmark & \cmark\,77{,}400 (167.9$\dagger$) & \xmark & \xmark \\
      & OpenRooms~\cite{li2021openrooms}           & Indoor  & 1{,}287 & — & — & \xmark & \cmark\,118{,}233 (91.9$\dagger$) & \cmark\,414 & \xmark \\
      & SpatialLM-Dataset~\cite{wang2025spatiallm} & Indoor  & 12{,}328 & — & — & \xmark & \xmark & \xmark & \cmark \\
    \midrule
    \multirow{2}{*}{Procedural}
      & ProcTHOR~\cite{deitke2022procthor}          & Indoor  & 10{,}000 & — & $\sim$747K$\dagger$ (74.7) & \xmark & \xmark & \xmark & \xmark \\
      & HSSD~\cite{khanna2024hssd}                  & Indoor  & 211 & — & $\sim$69{,}567$\dagger$ (329.7) & \xmark & \xmark & \xmark & \xmark \\
    \midrule
    \multirow{4}{*}{LLM / Agent}
      & Holodeck~\cite{yang2024holodeck}            & Indoor  & 120 & — & — (23.0) & \xmark & \xmark & \xmark & \cmark \\
      & SAGE~\cite{xia2026sage}                     & Indoor  & 10{,}000 & — & 565K (56.5$\dagger$) & \xmark & \xmark & \xmark & \cmark \\
      & MANSION~\cite{chi2026mansion}               & Indoor  & 1{,}000 & — & — & \xmark & \xmark & \xmark & \cmark \\
      & SceneSmith~\cite{pfaff2026scenesmith}       & Indoor  & 210 & — & $\sim$19.4K$\dagger$ (71.1/214.1) & \xmark & \xmark & \xmark & \cmark \\
    \midrule
    \textbf{Game Engine}
      & \textbf{Lumera-2K (Ours)}              & \textbf{In/Outdoor} & \textbf{2{,}513}
      & \textbf{3.73M (1{,}483)} & \textbf{63.00M (25{,}067)}
      & \textbf{\cmark\,102.6K (40.8)}
      & \textbf{\cmark\,95.1K (37.86)}
      & \textbf{\cmark\,2{,}513}
      & \textbf{\cmark} \\
    \bottomrule
  \end{tabular}}
\end{table*}

%% file: sections/dataset.tex
\section{Lumera-2K Dataset}
\label{sec:dataset}

Lumera-2K is a game-engine dataset for light-aware, editable scene parsing.
From $2{,}513$ UE5 projects collected from publicly available free assets on the internet, we extract object boxes, open-vocabulary object
labels, per-camera active parametric lights, SkyLight HDRI probes, and planned
camera views.  The dataset contains \textbf{$3.73$M components, $63$M object
instances, $102.6$K parametric lights, and $95.1$K camera views}, with per-scene
means of $1{,}483 / 25{,}067 / 40.8 / 37.86$.  Unlike common 3D scene datasets
~\cite{dai2017scannet,fu2021front,roberts2021hypersim,yang2024holodeck,wang2025spatiallm},
Lumera-2K provides countable engine-light annotations rather than only baked
illumination or global HDR signals.  A full dataset comparison appears in
Table~\ref{tab:dataset_complexity}; export fields and camera-planning details
are in Appendices~\ref{app:light_export} and~\ref{app:camera}.

\subsection{Scene-Aware Camera Planning}
\label{sec:dataset_camera}

The cameras shipped with UE5 projects are too sparse for supervision, averaging
only $3.0$ cameras per scene.  We therefore expand each project to $16$--$100$
informative views using a headless UE5 pipeline.  The planner clusters
foreground objects into interest regions, samples indoor cameras from room
segmentation and outdoor cameras from object/building anchors, rejects
wall-facing or near-empty views, and greedily selects a diverse set of cameras:
\begin{equation}
\label{eq:cam_score}
  s(v\mid S_t) \,=\, w_c\,\tfrac{|\mathcal{V}(v)\setminus\mathcal{C}_t|}{|\mathcal{F}|}
    \,+\, w_n\,N(v,S_t) \,+\, w_a\,A(v).
\end{equation}
Here $\mathcal{V}(v)$ is the foreground set visible from candidate view $v$,
$\mathcal{C}_t$ is accumulated coverage, $N(v,S_t)$ is position/orientation
novelty, and $A(v)$ is image fill.  A lightweight image-QA pass removes blank,
overexposed, underexposed, and low-content frames before high-resolution RGB,
G-buffer, and N3D export.

\subsection{Labels and Lights}
\label{sec:dataset_label_light}

Game assets often have internal names such as \texttt{SM\_chair\_01} or
\texttt{BP\_Seat\_A}.  We render isolated object views, combine them with weak
asset-name hints and one contextual scene image, and ask a VLM to return a
concise visual label.  The prompt requires image evidence to dominate project
naming, and a normalization pass suppresses shell labels such as wall, floor, and
editor placeholders.

For lights, we traverse PointLight, SpotLight, DirectionalLight, RectLight, and
SkyLight components at both scene and camera granularity.  The release keeps
native properties such as position, orientation, intensity units, attenuation,
SpotLight cone parameters, RGB color, and temperature.  In the supervised task
used here, each active non-SkyLight whose influence intersects the camera frustum
is abstracted to the seven-dimensional tuple in Eq.~\eqref{eq:light-tuple},
$(x,y,z,r,g,b,I)$, aligning light supervision with view-level object boxes while
preserving direct import into DCC tools.

%% file: sections/method.tex
\begin{figure*}[htbp]
  \centering
  \includegraphics[width=\textwidth]{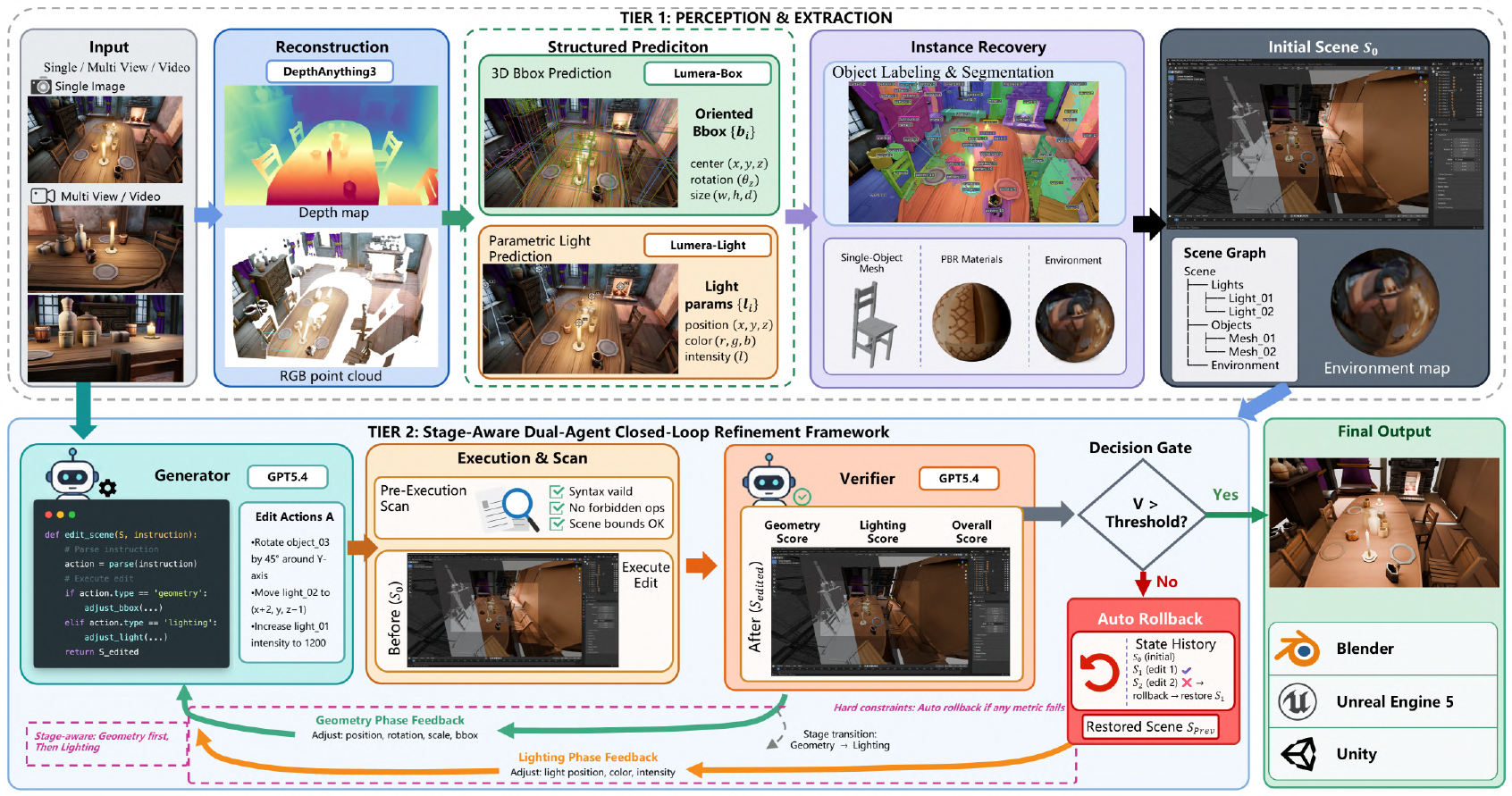}
  \caption{\textbf{Lumera pipeline.}
  \textbf{Top: structured parsing.}  A geometry front end reconstructs a colored
  point cloud from the input image; Lumera-Box and Lumera-Light, two
  independent SpatialLM SFT checkpoints, parse oriented object boxes and
  parametric light tuples $(x,y,z,r,g,b,I)$.  \textbf{Middle: instantiation and
  assembly.}  Boxes are projected back to the image to prompt instance masks;
  SAM3D reconstructs a textured mesh per object crop, IntrinsicHDR estimates an
  HDR environment probe, and a prior-driven bootstrap assembles these entities
  into an initial Blender scene $s_0$.  \textbf{Bottom: bounded refinement.}
  Generator and Verifier agents iterate in separate geometry and lighting
  stages.  The geometry stage can edit only object yaw and scale, while the
  lighting stage can edit only existing parametric lights, environment strength,
  and exposure.  Static scope checks, scene-difference validation, and rollback
  enforce these constraints.}
  \Description{Pipeline overview: geometry recovery, 3D box parsing, parametric
  light parsing, box-guided instance extraction, SAM3D object mesh recovery, HDR
  environment estimation, prior-driven scene assembly, and stage-aware
  constrained refinement.}
  \label{fig:pipeline}
\end{figure*}

\section{Method}
\label{sec:method}

\textbf{Design principle.}
Lumera treats game-scale image-to-scene reconstruction as
\emph{engine-native structured parsing followed by constrained reconstruction
and assembly}.  The output is not
an implicit radiance field or an unconstrained generated scene, but an explicit
engine scene made of object meshes with rigid transforms, parametric lights, an
HDR environment probe, and editable scene metadata.  This factorization avoids a
dense all-pairs instance model: boxes and lights are decoded as entity sequences,
object meshes are reconstructed independently from image crops, and the final
agentic loop is only allowed to edit a small whitelist of fields.

For the $v$-th planned camera view of a UE5 scene $S$, supervision is
\begin{equation}
  \label{eq:supervision-tuple}
  \mathcal{D}_v
  =(\mathcal{P}_v,\mathcal{Y}^{\text{txt}}_v,\mathcal{Y}^{\text{box}}_v,\mathcal{L}_v),
\end{equation}
where $\mathcal{P}_v\subset\mathbb{R}^3\times[0,255]^3$ is the colored point
cloud back-projected from rendered G-buffer depth,
$\mathcal{Y}^{\text{txt}}_v=\{l_i\}$ and
$\mathcal{Y}^{\text{box}}_v=\{b_i\}$ are the visible foreground labels and
oriented boxes, and $\mathcal{L}_v=\{\ell_j\}$ is the set of active engine lights
whose source or influence region intersects the view frustum.  We fine-tune two
independent SpatialLM-1.1-Qwen-0.5B~\cite{wang2025spatiallm} checkpoints,
collectively referred to as the \textbf{Lumera} parsers.  They share the base architecture,
point-cloud encoder, and location-token discretization $\Phi(\cdot)$, but use
separate data streams, task schemas, weights, and decoding settings.

\subsection{Lumera-Box: 3D Box Parsing}
\label{sec:method_bbox}

Lumera-Box keeps the SpatialLM architecture unchanged and changes the
supervision domain.  This matters because spatial LLMs trained mainly on
indoor scans and synthetic rooms transfer poorly to dense game-engine scenes
with outdoor extent, long-tail assets, and weak room-shell regularity.  We use
three practical adaptations during fine-tuning: a foreground vocabulary that
suppresses wall, floor, window, terrain, and editor-placeholder labels; multi-view
camera supervision from the same UE project; and class rebalancing that reduces
domination by frequent asset categories.

Each object instance is represented by an oriented 3D box
\begin{equation}
  \label{eq:bbox-tuple}
  b_i=(\mathbf{p}_i,\theta_i,\mathbf{s}_i)
  \in \mathbb{R}^3\times[-\pi,\pi)\times\mathbb{R}_{>0}^3,
\end{equation}
where $\mathbf{p}_i$ is the center, $\theta_i$ is yaw, and $\mathbf{s}_i$ is the
extent.  The serialized entity block is
$\tau^{\text{box}}_i=\Phi(l_i)\Vert\Phi(\mathbf{p}_i)\Vert\Phi(\theta_i)\Vert\Phi(\mathbf{s}_i)$.
Blocks are sorted with a deterministic $z$-major rule and wrapped in SpatialLM's
layout delimiters:
\begin{equation}
\label{eq:bbox-sequence}
  \mathcal{S}^{\text{box}}_v =
    \texttt{<|layout\_s|>}\;
    \biguplus_{i=1}^{N^{\text{box}}_v}\tau^{\text{box}}_i\;
    \texttt{<|layout\_e|>}.
\end{equation}
Under the instruction prefix \texttt{Detect boxes.} and the schema in
Appendix~\ref{app:schema}, Lumera-Box with parameters $\theta_{\text{box}}$
minimizes
\begin{equation}
\label{eq:loss-box}
  \mathcal{L}_{\text{box}} =
  -\mathbb{E}_{\mathcal{D}^{\text{box}}_{\text{train}}}
  \sum_t \log p_{\theta_{\text{box}}}(s_t^\star\mid
  s_{<t}^\star,\mathcal{P}_v,\mathcal{T}^{\text{box}}).
\end{equation}
At inference, we decode the sequence, invert $\Phi$, and recover
$(l_i,\mathbf{p}_i,\theta_i,\mathbf{s}_i)$ for each emitted instance.

\subsection{Lumera-Light: Parametric Light Parsing}
\label{sec:method_light}

Lumera-Light uses the same backbone and tokenization machinery, but a separate
light stream $\mathcal{D}^{\text{lit}}_{\text{train}}$, instruction prefix
\texttt{Detect lights.}, schema, weights $\theta_{\text{lit}}$, and decoding
parameters.  We decouple boxes and lights for three reasons.  First, boxes are
far denser than lights in most views, so joint training can bury the rare light
signal under high-frequency box tokens.  Second, independent checkpoints let
either task be improved or repaired without contaminating the other.  Third,
downstream assembly consumes boxes and lights as separate editable entity sets,
so joint decoding is not required.

The main paper predicts a compact, engine-importable light tuple
\begin{equation}
  \label{eq:light-tuple}
  \ell_j=(\mathbf{p}_j,\mathbf{c}_j,I_j)
  \in \mathbb{R}^3\times[0,255]^3\times\mathbb{R}_{\geq 0},
\end{equation}
where $\mathbf{p}_j$ is position, $\mathbf{c}_j$ is RGB color, and $I_j$ is
intensity.  Lumera-2K also stores native UE5 metadata such as type,
orientation, attenuation radius, cone angles, and temperature; these fields are
released and used by the adapter, but are not part of the first SFT target to
keep the prediction space stable.  The light entity block is
$\tau^{\text{lit}}_j=\Phi(\mathbf{p}_j)\Vert\Phi(\mathbf{c}_j)\Vert\Phi(I_j)$,
with target sequence
\begin{equation}
\label{eq:light-sequence}
  \mathcal{S}^{\text{lit}}_v =
    \texttt{<|layout\_s|>}\;
    \biguplus_{j=1}^{N^{\text{lit}}_v}\tau^{\text{lit}}_j\;
    \texttt{<|layout\_e|>}.
\end{equation}
Lumera-Light minimizes the analogous autoregressive objective
\begin{equation}
\label{eq:loss-lit}
  \mathcal{L}_{\text{lit}} =
  -\mathbb{E}_{\mathcal{D}^{\text{lit}}_{\text{train}}}
  \sum_t \log p_{\theta_{\text{lit}}}(s_t^\star\mid
  s_{<t}^\star,\mathcal{P}_v,\mathcal{T}^{\text{lit}}),
\end{equation}
and decoding returns $(\mathbf{p}_j,\mathbf{c}_j,I_j)$ for each predicted light.

\subsection{Mesh Recovery and Environment Lighting}
\label{sec:method_mesh}

Depth Anything 3~\cite{yang2025depthanything3} provides camera, depth, and a
colored point cloud for structured parsing.  This geometry is not treated as the
final scene; it acts as a metric bridge between the 3D entities and the input
image.  For each predicted box $\mathcal{B}_i$, we project the box into the
reference image to obtain a rectangle prompt $\hat{\mathbf{r}}_i$, combine it
with the predicted text label $l_i$, and run a SAM-family prompt-conditioned
segmenter:
\begin{equation}
  m_i=\textsc{Segment}(I_v;\hat{\mathbf{r}}_i,l_i).
\end{equation}
If the projected mask and the box inliers
$\mathcal{P}_i=\{p\in\mathcal{P}_v\mid p\in\mathcal{B}_i\}$ disagree because of
occlusion, calibration noise, or box drift, the 3D box remains the instance
identity and rigid boundary, while the mask only improves the RGB crop.  The
mesh module receives an alpha-matted crop with optional local context.

SAM3D~\cite{vorobyev2026sam3d} reconstructs a textured mesh from each crop; for
multi-view input we choose the least occluded representative view.  The mesh is
recovered in normalized object coordinates and placed back into the scene by
\begin{equation}
  \mathbf{T}_i =
  \mathbf{Trans}(\mathbf{p}_i)\mathbf{Rot}(\theta_i)\mathbf{Scale}(\mathbf{s}_i).
\end{equation}
This version focuses on foreground assets; large shells such as walls, floors,
ceilings, and terrain are handled by a separate shell pipeline.  Parametric
lights are complemented with global illumination: IntrinsicHDR
~\cite{dille2024intrinsichdr} estimates an HDR panorama or SkyLight cubemap from
the image, yielding the standard engine combination of explicit local lights
plus an HDR environment probe.

\subsection{Stage-Aware Agentic Refinement}
\label{sec:method_ric}

The initial scene can still contain yaw, scale, local placement, and lighting
residuals.  These errors are hard to fix with a single VLM call, and the full
pipeline is not differentiable through the renderer, parser, segmenter, and mesh
generator.  We therefore adapt the VIGA analysis-by-synthesis loop
~\cite{yin2026viga}, but make it stage-aware and executable.  The loop adds four
constraints: prior-driven initialization from parsed entities, separate geometry
and lighting stages, static and dynamic validation of every executed edit, and a
structured Verifier report whose fields map to the next Generator action.

The assembled scene is initialized as
$s_0=\textsc{Bootstrap}(\Pi)$ from
$\Pi=(\mathcal{B},\{\mathrm{Mesh}_i\},\mathcal{L}_0,\mathcal{H}_{\text{HDR}})$.
Here $\mathcal{L}_0$ can be predicted lights or a GT UE5 light export through
the adapter in Appendix~\ref{app:closed_loop}.  Generator $G$ and Verifier $V$
share a sliding conversation state
$M_0=\{\Theta^{\sigma_0},I_{\text{ref}},s_0\}$.  At iteration $t$ of stage
$\sigma$,
\begin{equation}
\label{eq:loop-step}
  (s_{t-1},M_t)\xrightarrow{G;\mathrm{exec}_\sigma;V}(s_t,M_{t+1}).
\end{equation}
$G$ proposes an action under the active scope, the constrained executor returns
a scene and violation set, $V$ produces a structured report, and the history is
updated with a sliding window.

Each stage has an edit scope
$\mathcal{E}_\sigma=(\mathcal{F}_\sigma,\mathcal{A}_\sigma)$, where
$\mathcal{F}_\sigma$ is the frozen entity set and $\mathcal{A}_\sigma$ is the
allowed change set:
\begin{align}
\label{eq:geom-allow}
\sigma=\text{geom}:&
\quad \mathcal{A}_\sigma=\{\Delta\theta_i,\Delta\mathbf{s}_i\},\quad
\Delta\mathbf{p}_i\equiv\mathbf{0},\\
\label{eq:light-allow}
\sigma=\text{light}:&
\quad \mathcal{A}_\sigma=\{\Delta\ell_j,\Delta E_{\text{env}},\Delta\gamma\}.
\end{align}
The geometry stage freezes cameras, lights, materials, mesh topology, object
identity, and object position by default; the lighting stage freezes cameras,
meshes, object transforms, topology, and materials.  The lighting stage can
modify only lights already present in $s_0$, environment strength, and exposure.
This split reduces the common failure mode where a VLM changes geometry while
trying to fix lighting, or changes lighting to compensate for missing geometry.

For a proposed action $a_t$, the executor applies three checks: a static scan for
out-of-scope code, a structured pre/post scene-difference check against the
field-level whitelist, and stage-specific precondition checks such as missing
light carriers.  Let $V_t$ be the union of these violations.  Execution is
\begin{equation}
\label{eq:exec-sigma}
  (s_t,V_t)=\mathrm{exec}_\sigma(s_{t-1},a_t):=
  \begin{cases}
    (\mathrm{exec}(s_{t-1},a_t),\emptyset), & V_t=\emptyset,\\
    (s_{t-1},V_t), & V_t\neq\emptyset.
  \end{cases}
\end{equation}
If a violation occurs, the framework restores the pre-execution snapshot and
writes the violation into the Verifier report.  The report contains free-text
visual differences, edit suggestions, a mirror of the current stage scope, any
execution violations, and two executable edit lists: $E^{\text{obj}}_t$ for
object yaw/scale repairs and $E^{\text{lit}}_t$ for light/environment/exposure
repairs.  Appendix~\ref{app:closed_loop} gives the full 13-field report schema,
action-space details, JSON repair layer, prompts, and pseudocode.

%% file: sections/experiments.tex
\section{Experiments}
\label{sec:exp}

\subsection{Setup and Protocol}
\label{sec:exp_impl}
\label{sec:exp_protocol}

Lumera-Box and Lumera-Light start from public
SpatialLM-1.1-Qwen-0.5B~\cite{wang2025spatiallm} weights and are fine-tuned on
disjoint box and light streams from Lumera-2K.  Splits are made at the UE
project level, yielding \textbf{val} ($1{,}316$ views) and \textbf{test}
($1{,}314$ views) subsets with both indoor and outdoor scenes.  The aligned box
benchmark compares \textbf{DetAny3D}, zero-shot \textbf{SpatialLM},
\textbf{N3D-VLM}, \textbf{WildDet3D}, and \textbf{Lumera-Box} under a
sanitized protocol that removes invalid labels, non-positive sizes, repeated
IDs, and ID fallback cases.  We report detection, metric geometry, semantics,
and structure metrics.  The light benchmark evaluates \textbf{Lumera-Light}
with position-first Hungarian matching at multiple thresholds; color and
intensity are measured on matched pairs.  Downstream assembly uses Depth
Anything 3~\cite{yang2025depthanything3}, a SAM-family segmenter,
SAM3D~\cite{vorobyev2026sam3d}, IntrinsicHDR~\cite{dille2024intrinsichdr}, and a
stage-aware Generator/Verifier loop with $T_g=T_l=12$ rounds.  Full
implementation details, threshold sweeps, and prompts are in the appendix.

\subsection{3D Box Parsing}
\label{sec:exp_bbox}

Table~\ref{tab:bbox_main} reports the main sanitized box benchmark on the merged
val+test split ($2{,}630$ views).  Lumera-Box is the strongest overall
method, leading on detection, metric geometry, semantics, and graph consistency.
WildDet3D remains stronger on SRF$\uparrow$ and anchor recall$\uparrow$,
indicating that relational anchor coverage is still a complementary weakness.

\begin{table*}[htbp]
  \centering\scriptsize
  \caption{Main 3D box parsing benchmark on the merged Lumera-2K val+test
  split under the sanitized protocol.  Arrows mark metric direction.}
  \label{tab:bbox_main}
  \setlength{\tabcolsep}{3pt}
  \resizebox{\textwidth}{!}{%
  \begin{tabular}{@{}lrrrrrrrrrrr@{}}
    \toprule
    Model
      & mAP$\uparrow$ & IoU-B$\uparrow$ & IoU-AABB$\uparrow$ & Chamf-L2$\downarrow$ & F-score$\uparrow$
      & C-MAE$\downarrow$ & Sem.$\uparrow$ & SRF$\uparrow$ & GCC$\uparrow$ & AWLE$\downarrow$ & Anc.R$\uparrow$ \\
    \midrule
    DetAny3D      & 0.0000 & 0.0004 & 0.0023 & 15.1763 & 0.0030 & 12.2640 & 0.0184 & 0.0084 & 0.2203 & 0.0854 & 0.0256 \\
    SpatialLM     & 0.0000 & 0.0030 & 0.0082 &  9.2587 & 0.0240 &  8.5801 & 0.0031 & 0.0018 & 0.2470 & \textbf{0.0127} & 0.0061 \\
    N3D-VLM       & 0.0015 & 0.0223 & 0.0286 & 3430.2389 & 0.0431 & 2188.3289 & 0.1451 & 0.0868 & 0.4375 & 48.1075 & 0.2139 \\
    WildDet3D     & 0.0021 & 0.0141 & 0.0144 &  6.4004 & 0.0566 &  7.0573 & 0.3181 & \textbf{0.5748} & 0.6127 & 0.4707 & \textbf{0.8811} \\
    \textbf{Lumera-Box} & \textbf{0.1141} & \textbf{0.2472} & \textbf{0.2734}
      & \textbf{3.4644} & \textbf{0.2762} & \textbf{3.9893}
      & \textbf{0.3827} & 0.4377 & \textbf{0.6676} & 0.1773 & 0.5607 \\
    \bottomrule
  \end{tabular}}
\end{table*}

Compared with WildDet3D, the strongest baseline by overall structure, Lumera-Box
improves mAP$\uparrow$ by $+0.1120$, IoU-B$\uparrow$ by $+0.2332$,
F-score$\uparrow$ by $+0.2196$, center MAE$\downarrow$ by $-3.0680$\,m, and
scene-semantic score$\uparrow$ by $+0.0645$.  N3D-VLM has the highest baseline
IoU$\uparrow$ but very large Chamfer$\downarrow$ and center errors$\downarrow$,
suggesting global translation or scale instability.  These numbers support the
main claim that game-domain supervision matters, while SRF$\uparrow$ and anchor
recall$\uparrow$ show that relation recovery is not solved.  
Fig.~\ref{fig:postref_bbox_first_page} visualizes the same trend against
SpatialLM, N3D-VLM, and WildDet3D.

\begin{table}[htbp]
  \centering
  \caption{Lumera-Box versus WildDet3D on the merged split.  $\Delta$ follows
  the metric direction, so positive means better for $\uparrow$ metrics and
  negative means better for $\downarrow$ metrics.}
  \label{tab:bbox_delta}
  \setlength{\tabcolsep}{3pt}
  \begin{tabular}{@{}lrrr@{}}
    \toprule
    Metric & Lumera-Box & WildDet3D & $\Delta$ \\
    \midrule
    mAP$\uparrow$        & 0.1141 & 0.0021 & $+0.1120$ \\
    IoU-B$\uparrow$      & 0.2472 & 0.0141 & $+0.2332$ \\
    IoU-AABB$\uparrow$   & 0.2734 & 0.0144 & $+0.2590$ \\
    Chamf-L2$\downarrow$ & 3.4644 & 6.4004 & $-2.9360$ \\
    F-score$\uparrow$    & 0.2762 & 0.0566 & $+0.2196$ \\
    C-MAE$\downarrow$    & 3.9893 & 7.0573 & $-3.0680$ \\
    Sem.$\uparrow$       & 0.3827 & 0.3181 & $+0.0645$ \\
    GCC$\uparrow$        & 0.6676 & 0.6127 & $+0.0549$ \\
    SRF$\uparrow$        & 0.4377 & 0.5748 & $-0.1371$ \\
    Anc.R$\uparrow$      & 0.5607 & 0.8811 & $-0.3204$ \\
    \bottomrule
  \end{tabular}
\end{table}

\subsection{Parametric Lights and Editable Assembly}
\label{sec:exp_light}

Table~\ref{tab:light_core} gives the required main-text light results on $575$
non-empty scenes ($284$ val and $291$ test).  Lumera-Light almost always
detects that a scene contains lights, but individual-light detection is still
hard: at $0.5$\,m, F1$\uparrow$ is $0.209$ with many false positives and misses.
Matched lights are localized within a median position error$\downarrow$
$0.261$\,m and have usable color estimates, but intensity remains coarse, with
log-scale MAE$\downarrow$ $0.431$ (about $2.7\times$ multiplicative error) and
Pearson $r\uparrow=0.628$.

\begin{table}[htbp]
  \centering
  \caption{Lumera-Light on 575 non-empty scenes.  Matching is position-first at
  0.5\,m unless stated.}
  \label{tab:light_core}
  \setlength{\tabcolsep}{3pt}
  \begin{tabular}{@{}lr@{}}
    \toprule
    Metric & Value \\
    \midrule
    Nonempty scene recall$\uparrow$ & 0.998 \\
    Count MAE$\downarrow$ / exact count$\uparrow$ & 2.30 / 0.442 \\
    P$\uparrow$ / R$\uparrow$ / F1$\uparrow$ @ 0.5\,m & 0.218 / 0.201 / 0.209 \\
    Joint success rate$\uparrow$ & 0.099 \\
    XYZ mean / median / P90 (m)$\downarrow$ & 0.263 / 0.261 / 0.295 \\
    $\Delta E_{2000}$ mean / median$\downarrow$ & 4.98 / 4.59 \\
    Intensity log10 MAE$\downarrow$ / Pearson $r\uparrow$ & 0.431 / 0.628 \\
    Pairwise dist. consistency$\uparrow$ & 0.901 \\
    \bottomrule
  \end{tabular}
\end{table}

Table~\ref{tab:light_thresh_main} shows the localization-threshold sweep.  A
looser threshold improves F1$\uparrow$ from $0.209$ at $0.5$\,m to $0.456$ at
$2.0$\,m, which means the model often predicts the rough light neighborhood but
does not localize enough sources precisely.  Formatting is not the limiting
factor: among $2{,}536$ accepted predicted lights, only two have invalid RGB
values, and we observe no regex mismatches, non-finite values, duplicate IDs, or
ID regressions.  The bottleneck is semantic and geometric recall of individual
sources, especially off-screen or near-camera lights whose influence is visible
but source geometry is ambiguous.  The full color/intensity breakdown is in
Appendix~\ref{app:light}.
Fig. ~\ref{fig:postref_light_first_page} provides the
corresponding post-reference qualitative audit, showing our inferred parametric
lights against GT light layouts.

\begin{table}[htbp]
  \centering\scriptsize
  \caption{Light detection sweep over position thresholds.  Color
  $\Delta E_{2000}\downarrow$ is measured on matched lights.}
  \label{tab:light_thresh_main}
  \setlength{\tabcolsep}{2pt}
  \resizebox{\columnwidth}{!}{%
  \begin{tabular}{@{}rrrrrrrrrr@{}}
    \toprule
    Thr. & TP$\uparrow$ & FP$\downarrow$ & FN$\downarrow$ & P$\uparrow$ & R$\uparrow$ & F1$\uparrow$ & JSR$\uparrow$ & Dist.$\downarrow$ & $\Delta E\downarrow$ \\
    \midrule
    0.25 &  289 & 2247 & 2459 & 0.114 & 0.105 & 0.109 & 0.051 & 0.164 & 4.45 \\
    0.50 &  553 & 1983 & 2195 & 0.218 & 0.201 & 0.209 & 0.099 & 0.266 & 5.72 \\
    1.00 &  885 & 1651 & 1863 & 0.349 & 0.322 & 0.335 & 0.168 & 0.441 & 6.63 \\
    2.00 & 1205 & 1331 & 1543 & 0.475 & 0.439 & 0.456 & 0.248 & 0.727 & 6.85 \\
    \bottomrule
  \end{tabular}}
\end{table}

\textbf{Editable assembly.}
On a 55-instance indoor scene driven by GT boxes, the bounded refinement loop
converges within 12 geometry rounds: a VLM score$\uparrow$ improves from $6.1$
to $8.3$, and Chamfer distance$\downarrow$ drops to $71.8\%$ of the baseline.
On outdoor scenes with
poor initial boxes, the loop provides little benefit, confirming that agentic
refinement is an editor over a usable parse rather than a replacement for
structured parsing.  A GT-driven lighting ablation is shown in
Appendix~\ref{app:lighting_ablation}; the post-reference figure pages further
compare full reconstruction with CAST, MIDI, SceneGen, and VIGA.

\begin{figure}[htbp]
  \centering
  \includegraphics[width=\linewidth]{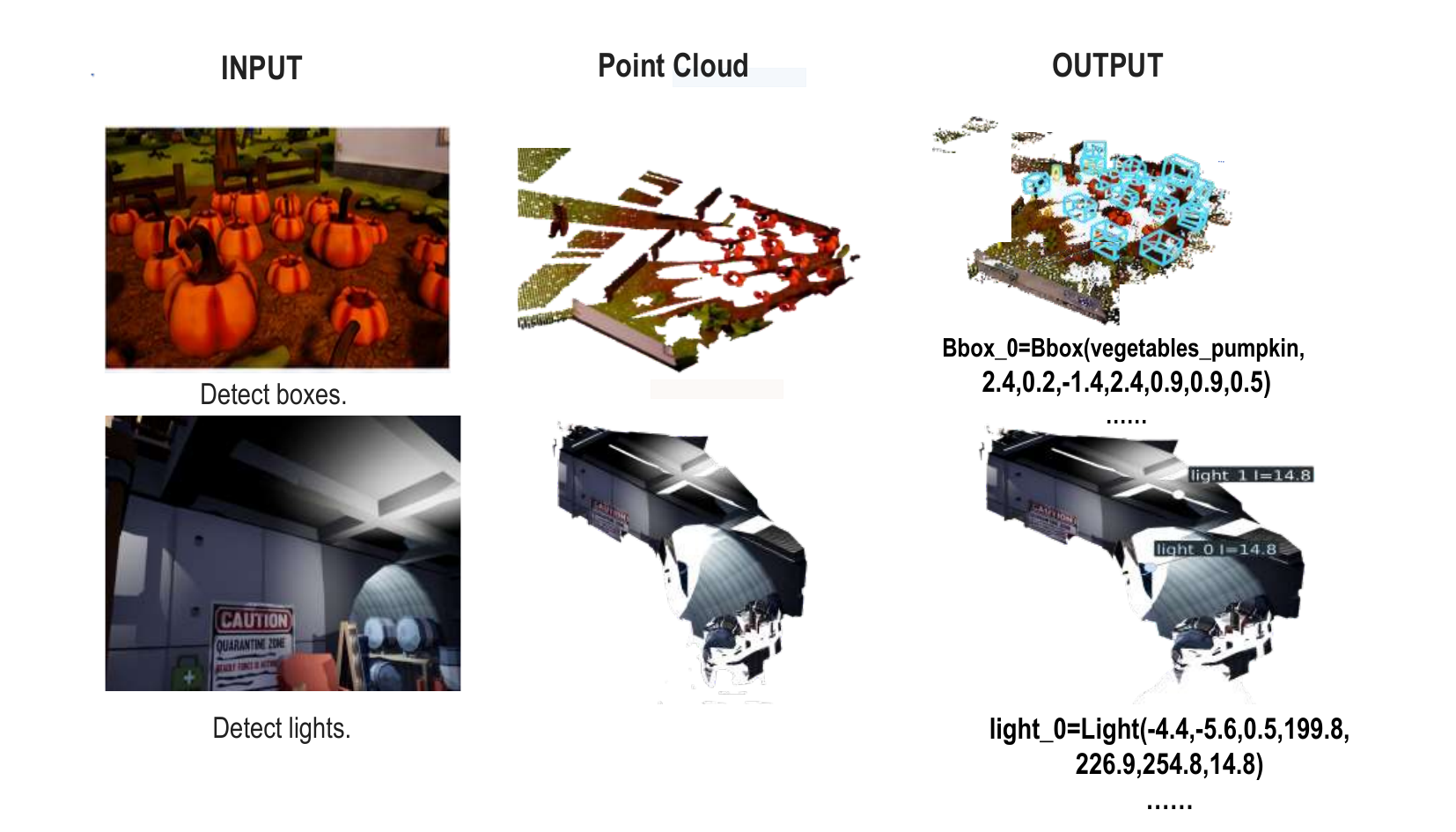}
  \caption{\textbf{Schema-conditioned 3D scene parsing for object boxes and lights.}
  \textbf{Top: box parsing.} A colored point cloud is paired with a task prompt
  and a \texttt{Bbox} schema, enabling the model to generate structured object
  layouts of the form \texttt{Bbox(class, x, y, z, yaw, sx, sy, sz)}. This
  represents category-aware, orientation-aware 3D object localization as
  autoregressive layout-code prediction. \textbf{Bottom: light parsing.} Using
  the same point-cloud-conditioned generation framework, the model is instead
  prompted with a \texttt{Light} schema and predicts parametric light tuples
  \texttt{Light(x, y, z, r, g, b, I)}, capturing light position, color, and
  intensity. The figure highlights that box detection and light estimation are
  not modeled as separate detection heads, but as two schema-specific instances
  of a unified SpatialLM-style 3D-to-code generation task.}
  \Description{Unified SpatialLM-style formulation for 3D bounding box parsing
  and light source parsing from point clouds. The upper branch predicts object
  boxes as structured code, while the lower branch predicts parametric lights as
  structured code.}
\end{figure}

\begin{figure*}[htbp]
  \centering
  \includegraphics[width=\linewidth]{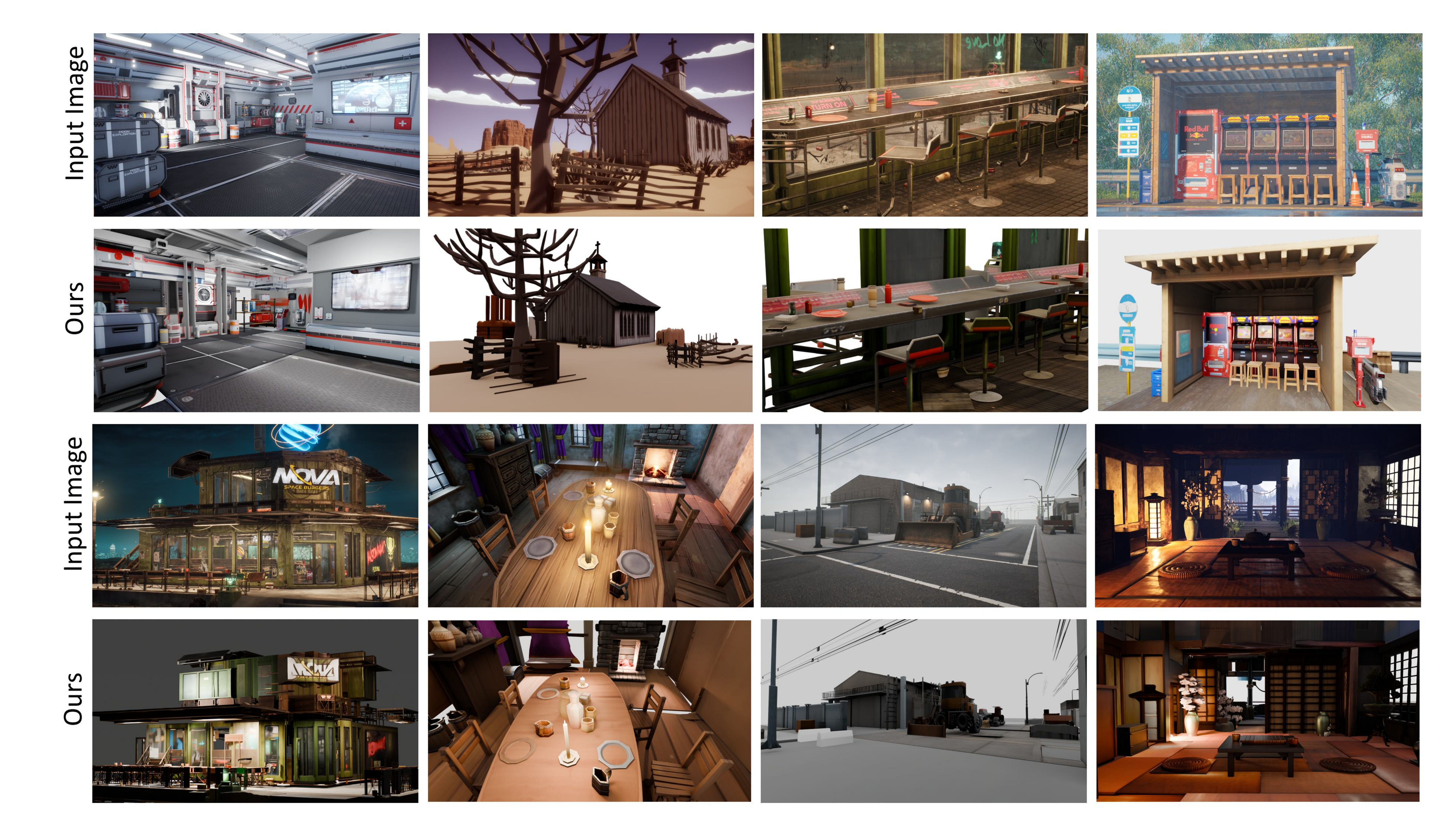}
    \caption{\textbf{Case studies of image-conditioned 3D scene generation.}
    Given a single input image, our method reconstructs coherent 3D scenes with plausible object layouts, geometry, and lighting across diverse environments.}
\end{figure*}

%% file: sections/conclusion.tex
\section{Conclusion}
\label{sec:conclusion}

We introduced Lumera, a benchmark and reference pipeline for engine-native,
light-aware 3D scene parsing, reconstruction, and assembly from a single image.
The key contribution is to
bring engine-native parametric lights into the editable-scene reconstruction
problem: Lumera-2K records them as structured entities, Lumera-Light
predicts them from scene point clouds, and the assembly pipeline keeps them
editable in Blender or UE5.  The current system still has clear limitations in
strict 3D box accuracy, yaw estimation, light recall for off-frustum or
near-camera sources, and generalization beyond UE-style assets
(Appendix~\ref{app:limitations}).  These limitations are precisely why a
light-aware dataset and benchmark are useful: they expose the unsolved parts of
game-scale editable reconstruction under a representation that downstream tools
can actually use.

%% file: sections/postref_figures.tex
\IfFileExists{img/light_qual.pdf}{%
  \begin{figure*}[p]
    \centering
    \includegraphics[width=\textwidth,height=\textheight,keepaspectratio]{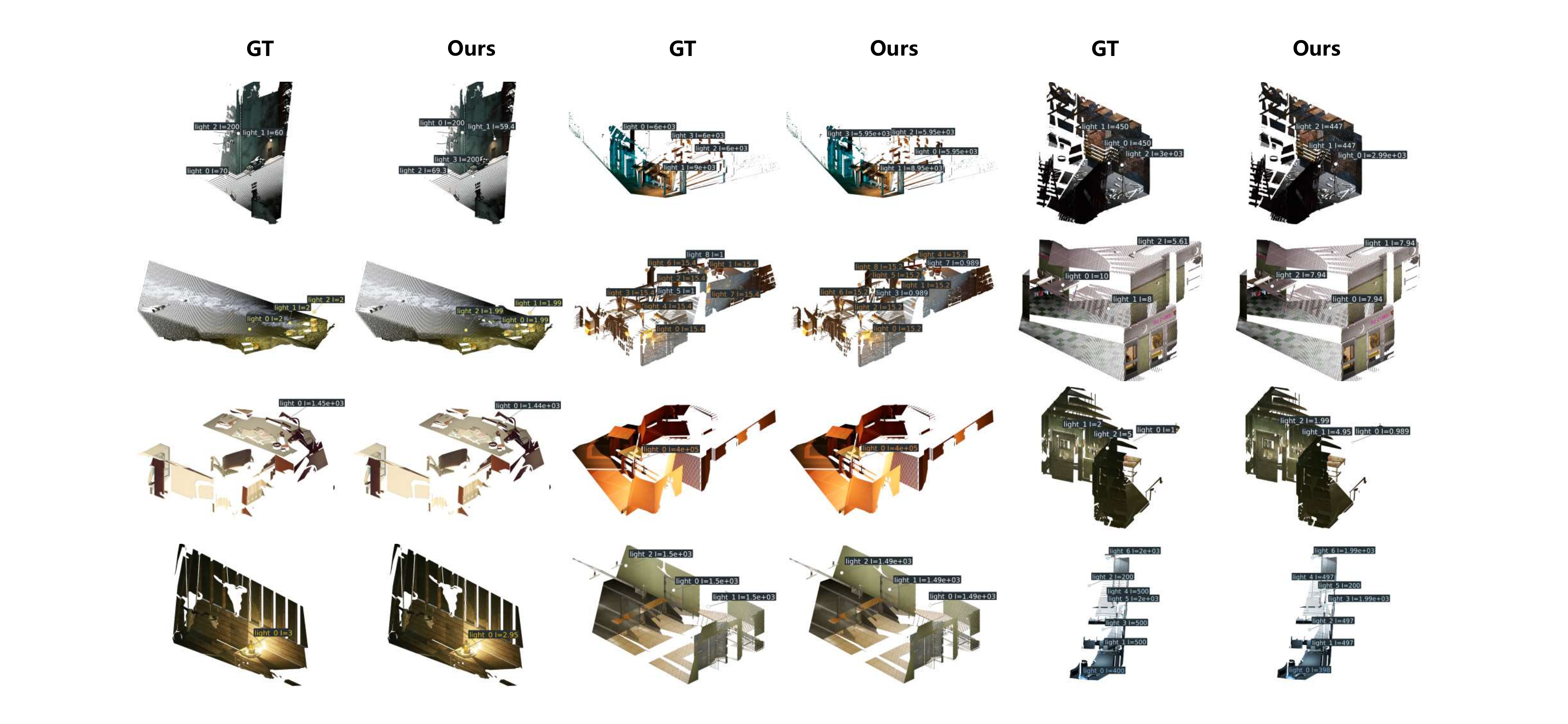}
    \caption{\textbf{Qualitative inference results from Lumera-Light.}
    Lumera-Light predicts engine-native parametric lights, shown against
    ground-truth light layouts in game-scale point clouds.}
    \label{fig:postref_light_first_page}
    \Description{Post-reference qualitative figure page showing parametric
    light prediction results from Lumera-Light.}
  \end{figure*}
}{}

\IfFileExists{img/box_qual.pdf}{%
  \begin{figure*}[p]
    \centering
    \includegraphics[width=\textwidth,height=0.9\textheight,keepaspectratio]{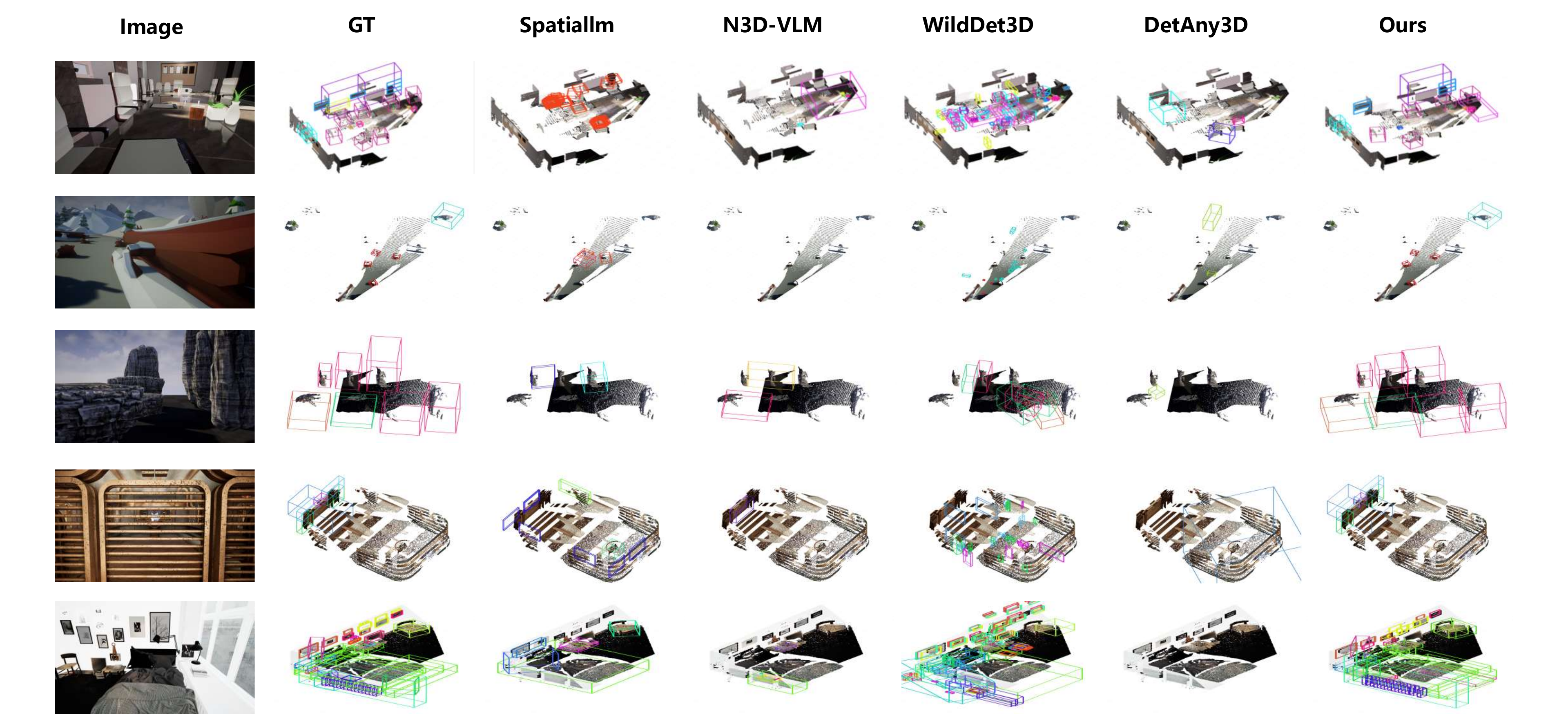}
    \caption{\textbf{Qualitative inference results from Lumera-Box.}
    3D box parsing comparisons against SpatialLM, N3D-VLM, and WildDet3D,
    where Lumera-Box recovers denser and more spatially aligned editable
    object layouts.}
    \label{fig:postref_bbox_first_page}
    \Description{Post-reference qualitative figure page showing oriented 3D
    bounding-box prediction results from Lumera-Box.}
  \end{figure*}
}{}

\IfFileExists{img/scene_qual.pdf}{%
  \begin{figure*}[p]
    \centering
    \includegraphics[width=\textwidth,height=0.9\textheight,keepaspectratio]{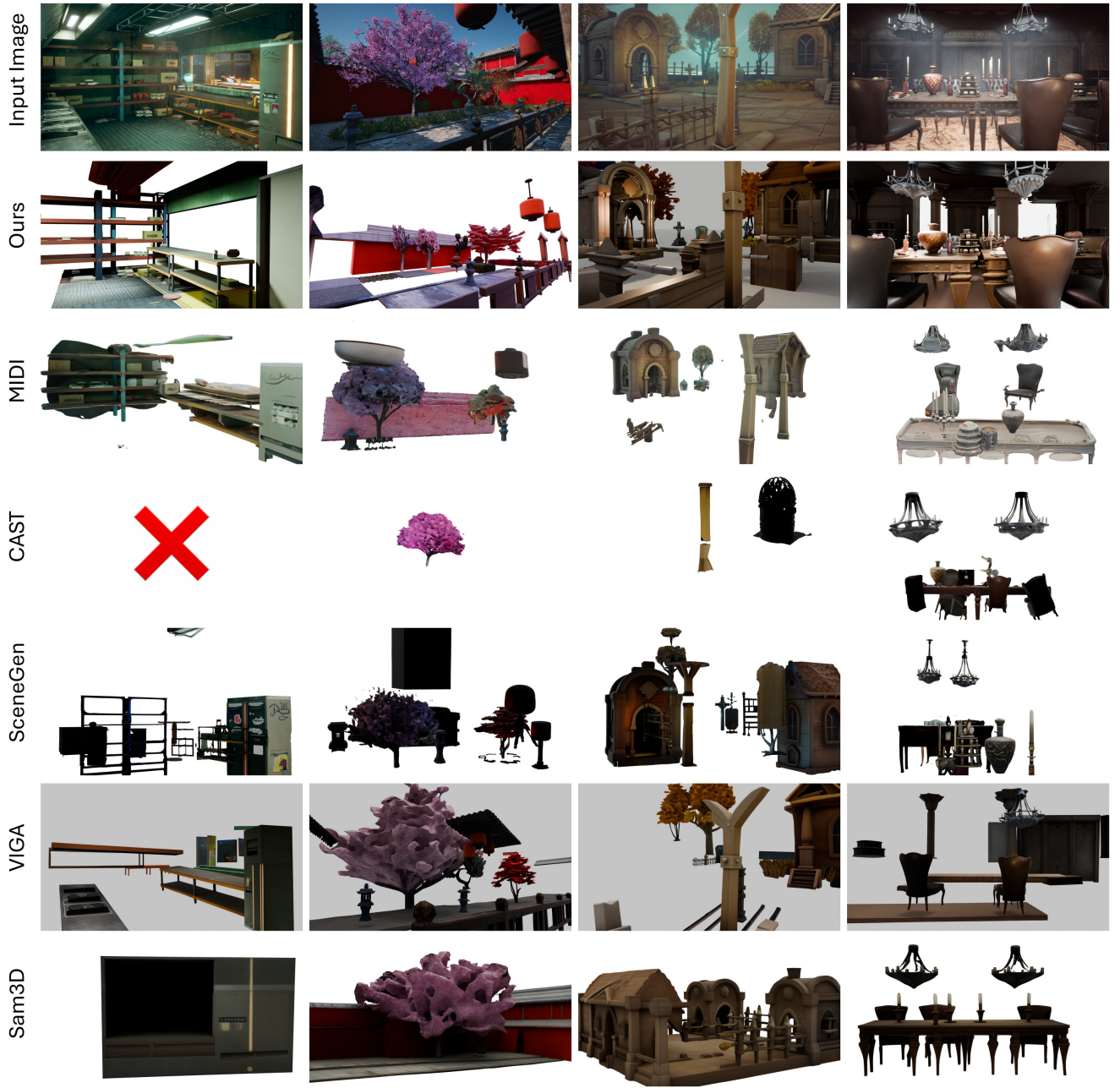}
    \caption{\textbf{Qualitative comparison on image-conditioned 3D scene generation.}
    Each column shows one input image and the corresponding 3D scene generated by different methods, including Ours, MIDI, CAST, SceneGen, VIGA, and Sam3D. Compared with prior approaches, our method better preserves the scene-level layout, object composition, and spatial relationships of the input image. It reconstructs more complete and coherent 3D scenes across both indoor and outdoor environments, while competing methods often produce incomplete geometry, isolated objects, missing structures, or inconsistent arrangements.}
    \Description{Post-reference qualitative figure page showing oriented 3D
    bounding-box prediction results from Lumera-Box.}
  \end{figure*}
}{}

%% file: sections/appendix.tex
\section{Camera Planning and UE5 Rendering Details}
\label{app:camera}

This appendix expands the camera-planning and rendering pipeline summarized in
Sec.~\ref{sec:dataset_camera}.  The pipeline has five stages, and all numeric
values are the defaults in our release implementation.

\subsection{Stage 1: Foreground Filtering and XY Clustering}
We traverse all static-mesh actors with the engine-side Python API.  Structural
background actors are filtered by tag substrings, including walls, floors,
ceilings, doors, windows, stairs, roofs, terrain, grass, soil, and sky.  We also
remove tiny debris ($<18$\,cm by AABB extent) and very large shell objects
($>4$\,m).  The remaining foreground objects are clustered on a $2.2$\,m XY
grid.  Each cluster roughly corresponds to a room or an outdoor region of
interest.  We cap the number of clusters at 32 and keep the most diverse subset.

\subsection{Stage 2: Indoor Room Segmentation and Multi-Mode Candidate Sampling}
At camera height, roughly $1.6$\,m, we run a 2D flood-fill room segmentation.
Wall and doorway geometry is projected to a 40\,cm occupancy grid, connected
free-space components are extracted, and each foreground cluster is assigned to
the room that contains it.  Candidate cameras are then generated with three
priority levels: (i) \emph{in-room grid} sampling, which uniformly samples
locations inside the segmented room while keeping at least 60\,cm margin from
walls; (ii) \emph{free-interior fallback}, which samples inside the cluster AABB
when flood-fill fails or the room is too small; and (iii) \emph{orbit mode},
used when no reliable room information is available, e.g., outdoors, with radii
$0.9$ and $1.25$ times the cluster diameter and six azimuth samples.  All
candidate locations must satisfy body and camera clearance constraints
($\geq 25$ / $20$\,cm, checked by engine ray casts) and lie inside the scene
shell.  We then expand each location over FOV $[45^\circ,95^\circ]$, six yaw
templates, and a mild downward pitch.

\subsection{Stage 3: View-Level Occlusion and Visibility Filtering}
For each candidate view $v$, we perform two geometric tests.  First, a
\emph{wall-occlusion test} intersects the 2D segment from the camera to each
foreground-object center with all wall AABBs, removing blocked objects from the
candidate visible set.  Second, a \emph{near-structure rejection test} samples
nine representative frustum rays; if most hit a nearby structural surface
($\leq 2.4$\,m) before any foreground object, the candidate is rejected as a
wall-facing view.  Accepted candidates must see at least three foreground
objects, with at least one projected object box covering $\geq 0.45\%$ of the
image area.

\subsection{Stage 4: Greedy Coverage with Diversity Constraints}
The score is Eq.~\eqref{eq:cam_score}.  Position/orientation NMS requires any
two selected cameras to be at least $4$\,m apart and differ by at least
$12^\circ$ in yaw.  Selection stops when total coverage reaches $0.99$, marginal
coverage falls below $0.008$, or the selected count reaches the hard cap of
$150$.  We then allow each cluster to add up to six high-fill-ratio views.  The
indoor and outdoor planners share this scoring interface and the downstream QA
and rendering pipeline.  Their anchor and sampling differences are summarized in
Table~\ref{tab:cam_planner}.

\subsection{Stage 5: Image-Level QA Filtering}
Even after geometric filtering, candidate cameras may produce uninformative
frames, such as views pressed against a wall, near-solid-color images, or frames
where foreground objects are almost fully occluded.  For each preview image
resized to $160\!\times\!90$, we compute lightweight statistics: global luminance
mean and standard deviation $\sigma_L$, Sobel edge density $e$, and dominant
color ratio $\rho_c$ from an 8-bin-per-channel color histogram.  We also compute
the same statistics in the central $80\!\times\!46$ crop,
$\sigma_L^c, e^c, \rho_{cc}$.  Threshold rules detect five failure modes:
(i) globally too dark or too bright blank frames, (ii) large solid-color frames,
(iii) center-only solid-color frames with structure in the corners,
(iv) globally low-detail frames, and (v) parse or render failures.  Passing
views receive a quality score
\begin{equation}
\label{eq:qa_score}
  q \,=\, 0.34\,\widetilde{\sigma_L} + 0.34\,\widetilde{e}
        \,+\, 0.18\,(1\!-\!\rho_c) + 0.14\,(1\!-\!\rho_{cc}),
\end{equation}
where $\widetilde{\sigma_L}$ and $\widetilde{e}$ are normalized by reference
scales.  To avoid discarding all views for small scenes, recoverable categories
such as solid-color and low-content frames are sorted by quality and partially
reintroduced to satisfy a minimum view count and retention ratio.  Table
~\ref{tab:cam_qa} reports rejection reasons on 200 sampled scenes.

\subsection{Linux Headless UE5 Rendering}
To reduce process-start overhead, we use a batched backend: a persistent Unreal
process switches the render target through the $K$ selected cameras of a scene.
After each switch, the renderer waits several frames for auto exposure,
streaming, Lumen global illumination, and material compilation to stabilize
before capturing high-resolution screenshots; a global warm-up is inserted
before the first frame.  To handle occasional GPU interruptions, cameras are
chunked with short gaps between chunks, and the pipeline falls back to a
one-camera-per-process mode when needed.

Cross-scene tone consistency is controlled by a stable rendering preset: Lumen
diffuse global illumination is kept, while motion blur, depth of field, lens
flare, Lumen reflections, screen-space reflections, and bloom are disabled.  We
also fix tone-mapping quality and exposure histogram settings and use an
exposure bias to suppress extreme brightness.  For each QA-passing camera, we
export a G-buffer probe containing scene depth, world normal, base color,
metallic/roughness, and final RGB, then export a multi-view N3D bundle using the
probe RGB/depth as coverage signals.  When memory pressure is high, each export
batch is reduced or degraded to single-camera mode.

\begin{table}[H]
  \centering\footnotesize
  \caption{Preview-QA rejection reasons on 200 sampled Lumera-2K scenes.}
  \label{tab:cam_qa}
  \setlength{\tabcolsep}{4pt}
  \resizebox{\columnwidth}{!}{%
  \begin{tabular}{@{}lcc@{}}
    \toprule
    Rejection reason & Share among rejected & Share among all candidates \\
    \midrule
    Large solid-color region     & 38.4\% & 9.7\% \\
    Center-only solid color      & 27.1\% & 6.9\% \\
    Globally low content         & 14.6\% & 3.7\% \\
    Too dark / too bright blank  & 11.2\% & 2.8\% \\
    Render failure / missing     &  8.7\% & 2.2\% \\
    \midrule
    \textbf{Total}               & 100.0\% & \textbf{25.3\%} \\
    \bottomrule
  \end{tabular}}
\end{table}

\begin{table}[H]
  \centering\footnotesize
  \caption{Key differences between the indoor and outdoor camera planners.}
  \label{tab:cam_planner}
  \setlength{\tabcolsep}{3pt}
  \resizebox{\columnwidth}{!}{%
  \begin{tabular}{@{}p{1.6cm}p{2.9cm}p{2.9cm}@{}}
    \toprule
    Module & Indoor planner & Outdoor planner \\
    \midrule
    Anchor       & Room flood-fill $+$ cluster & Building/object cluster $+$ coarse ground trace \\
    Candidates   & In-room grid / free interior / orbit & Orbit $+$ facade side rays $+$ grid \\
    Height       & Grounded $+$ 160\,cm eye height & Multi-level $\{1.6, 6.0, 18.0\}$\,m \\
    NavMesh dep. & \xmark & \xmark \\
    FOV range    & $45$--$95^{\circ}$ & $42$--$88^{\circ}$ \\
    Typical cov. & $\geq 0.99$ & $\geq 0.75$ \\
    \bottomrule
  \end{tabular}}
\end{table}

\section{Schemas for the Two Independent SpatialLM SFT Tasks}
\label{app:schema}

Equations~\eqref{eq:bbox-sequence} and~\eqref{eq:light-sequence} define the
target token sequences for Lumera-Box and Lumera-Light.  Each task
instruction, $\mathcal{T}^{\text{box}}$ or $\mathcal{T}^{\text{lit}}$, declares a
fixed field order and field type.  Table~\ref{tab:dual_schema} lists the two
schemas.  Numeric fields are declared as integers to guide SpatialLM's
location-token discretization $\Phi(\cdot)$; the continuous values in
Eqs.~\eqref{eq:bbox-tuple}--\eqref{eq:light-tuple} are recovered by
$\Phi^{-1}$.

\begin{table}[H]
  \centering\footnotesize
  \caption{Schema declarations for the two independent SpatialLM SFT tasks.
  Field order is fixed.  Numeric fields are represented through location-token
  discretization $\Phi(\cdot)$.}
  \label{tab:dual_schema}
  \setlength{\tabcolsep}{6pt}
  \begin{tabular}{@{}ll@{}}
    \toprule
    BBox schema & Light schema \\
    \midrule
    class (str)   & position\_x \\
    position\_x   & position\_y \\
    position\_y   & position\_z \\
    position\_z   & color\_r    \\
    angle\_z      & color\_g    \\
    scale\_x      & color\_b    \\
    scale\_y      & intensity   \\
    scale\_z      &             \\
    \bottomrule
  \end{tabular}
\end{table}

\section{Full List of Engine-Native Light Parameters}
\label{app:light_export}

The main text uses only position, color, and intensity as the SpatialLM
supervision target.  Lumera-2K exports five groups of native attributes for
every light and releases all of them:

\begin{itemize}[leftmargin=*, itemsep=2pt]
  \item \textbf{State and frustum relation}: light identifier, type, active
    state, whether the light source is visible in the current camera, and
    whether its influence region intersects the frustum.
  \item \textbf{Position and orientation}: world position in centimeters, world
    rotation, and camera-relative position $(x,y,z)$ in centimeters.
  \item \textbf{Intensity and attenuation}: UE intensity, intensity unit
    (lumen or candela, for cross-engine conversion), and maximum influence
    radius.
  \item \textbf{SpotLight shape}: inner and outer cone angles, source radius,
    source length, source width, and source height.
  \item \textbf{Color and temperature}: RGB color, Kelvin color temperature, and
    a Boolean indicating whether color temperature is enabled.
\end{itemize}

\noindent
We parameterize only $(x,y,z,r,g,b,I)$ in the SpatialLM output to keep training
stable.  The remaining fields are used as controllable metadata in the initial
UE-to-Blender light adapter and in the light-edit interface, and they enable
future work on finer-grained light prediction.

\section{Supplementary Metrics for Box Evaluation}
\label{app:bbox_geo}

Table~\ref{tab:bbox_main} in the main text reports the merged val+test benchmark.
This appendix gives the corresponding per-split results and the improvement over
the previous \textit{ckpt5000} box model.

\noindent
\textbf{Interpretation.}
Lumera-Box is stable across val and test: its mAP$\uparrow$ changes from
$0.1237$ to $0.1227$, IoU-B$\uparrow$ from $0.2482$ to $0.2463$, and
scene-semantic score$\uparrow$ from $0.3806$ to $0.3847$.  The relative ordering
of the baselines is also stable: WildDet3D consistently dominates SRF$\uparrow$
and anchor recall$\uparrow$, while N3D-VLM remains the strongest baseline in
IoU-B$\uparrow$/IoU-AABB$\uparrow$ but suffers from very large global geometry
errors$\downarrow$.

\begin{table}[H]
  \centering\footnotesize
  \caption{Progress from the previous \textit{ckpt5000} box model to the
  reported Lumera-Box checkpoint on the merged val+test split.}
  \label{tab:bbox_format}
  \setlength{\tabcolsep}{6pt}
  \begin{tabular}{@{}lrrr@{}}
    \toprule
    Metric & ckpt5000 & Lumera-Box & Improvement$\uparrow$ \\
    \midrule
    mAP$\uparrow$              & 0.0333 & 0.1141 & $+0.0808$ \\
    IoU-B$\uparrow$            & 0.1259 & 0.2472 & $+0.1213$ \\
    F-score$\uparrow$          & 0.1624 & 0.2762 & $+0.1138$ \\
    Scene semantic$\uparrow$   & 0.2343 & 0.3827 & $+0.1484$ \\
    SRF$\uparrow$              & 0.2774 & 0.4377 & $+0.1603$ \\
    GCC$\uparrow$              & 0.5696 & 0.6676 & $+0.0980$ \\
    Anchor recall$\uparrow$    & 0.3769 & 0.5607 & $+0.1838$ \\
    \bottomrule
  \end{tabular}
\end{table}

\noindent
The new checkpoint improves all reported detection, geometry, semantic, and
structure metrics over the previous model.  This is why the main text reports
the aligned \textit{clean\_move1} benchmark rather than the earlier ckpt5000
numbers.

\section{Complete Light-Prediction Metrics}
\label{app:light}

Table~\ref{tab:light_main} reports the light-prediction metrics used in
Sec.~\ref{sec:exp_light}.  The evaluation covers $575$ non-empty scenes
($284$ val and $291$ test) and uses position-first Hungarian matching, followed
by color and intensity evaluation on matched pairs.  Table~\ref{tab:light_thresh}
gives a sweep over position thresholds.

\begin{table}[H]
  \centering\footnotesize
  \caption{Main light-prediction metrics under the sanitized protocol, using a
  0.5\,m threshold unless otherwise stated.  Arrows mark metric direction.  The
  benchmark contains no empty GT scenes, so empty-scene accuracy is not
  applicable.}
  \label{tab:light_main}
  \setlength{\tabcolsep}{4pt}
  \begin{tabular}{@{}llr@{}}
    \toprule
    Group & Metric & Value \\
    \midrule
    \multirow{3}{*}{Scene existence}
      & num\_common\_files            & 575 \\
      & nonempty\_scene\_recall$\uparrow$ & 0.998 \\
      & catastrophic\_miss\_rate$\downarrow$ & 0.002 \\
    \midrule
    \multirow{4}{*}{Count alignment}
      & exact\_count\_accuracy$\uparrow$ & 0.442 \\
      & count\_mae$\downarrow$        & 2.30 \\
      & over\_pred\_rate$\downarrow$  & 0.252 \\
      & under\_pred\_rate$\downarrow$ & 0.306 \\
    \midrule
    \multirow{3}{*}{Detection @ 0.5\,m}
      & precision$\uparrow$           & 0.218 \\
      & recall$\uparrow$              & 0.201 \\
      & F1$\uparrow$                  & 0.209 \\
    \midrule
    \multirow{2}{*}{Joint metrics}
      & joint\_success\_rate$\uparrow$ & 0.099 \\
      & valid matched pairs           & 553 \\
    \midrule
    \multirow{3}{*}{Geometry (m)}
      & xyz mean / median$\downarrow$ & 0.263 / \textbf{0.261} \\
      & xyz $P_{90}\downarrow$        & 0.295 \\
      & chamfer / EMD$\downarrow$     & 7.108 / 6.465 \\
    \midrule
    \multirow{3}{*}{Color}
      & $\Delta E_{2000}$ mean / median$\downarrow$ & 4.98 / \textbf{4.59} \\
      & RGB MAE / RMSE$\downarrow$                  & 12.68 / 17.17 \\
      & luminance MAE / chroma err.$\downarrow$     & 11.36 / 7.75 \\
    \midrule
    \multirow{3}{*}{Intensity}
      & log10 MAE$\downarrow$           & \textbf{0.431} \\
      & linear MAE$\downarrow$          & 1486.3 \\
      & Pearson $r\uparrow$             & \textbf{0.628} \\
    \midrule
    \multirow{3}{*}{Scene structure}
      & pairwise dist. consistency$\uparrow$ & 0.901 \\
      & height dist. distance$\downarrow$   & 0.093 \\
      & nn spacing error$\downarrow$        & 0.139 \\
    \bottomrule
  \end{tabular}
\end{table}

\begin{table}[H]
  \centering\footnotesize
  \caption{Light detection under different position thresholds.  Arrows mark
  metric direction.  Color $\Delta E_{2000}$ is measured only on matched lights.
  Even at $2$\,m, F1$\uparrow$ is $0.456$, so individual-light recall and false
  positives remain the main bottlenecks.}
  \label{tab:light_thresh}
  \setlength{\tabcolsep}{3pt}
  \resizebox{\columnwidth}{!}{%
  \begin{tabular}{@{}rrrrrrrrr@{}}
    \toprule
    xyz thr (m) & TP$\uparrow$ & FP$\downarrow$ & FN$\downarrow$ & P$\uparrow$ & R$\uparrow$ & F1$\uparrow$ & JSR$\uparrow$ & mean dist$\downarrow$ \\
    \midrule
    0.25 &  289 & 2247 & 2459 & 0.114 & 0.105 & 0.109 & 0.051 & 0.164 \\
    0.50 &  553 & 1983 & 2195 & 0.218 & 0.201 & 0.209 & 0.099 & 0.266 \\
    1.00 &  885 & 1651 & 1863 & 0.349 & 0.322 & 0.335 & 0.168 & 0.441 \\
    2.00 & 1205 & 1331 & 1543 & 0.475 & 0.439 & 0.456 & 0.248 & 0.727 \\
    \bottomrule
  \end{tabular}}
\end{table}

\noindent
\textbf{Intensity metrics.}
We evaluate luminous intensity on the $553$ lights matched at the $0.5$\,m
threshold.  The primary metric is log-space absolute error,
\[
  \frac{1}{N}\sum_i \left|\log_{10} I_i^{\text{pred}}
  - \log_{10} I_i^{\text{gt}}\right|,
\]
because perceived brightness is closer to logarithmic than linear.  The value
$0.431$ corresponds to an average multiplicative error of
$10^{0.431}\!\approx\!2.7\times$.  Pearson correlation $r\uparrow=0.628$ shows
that the model has a moderate ability to rank brighter and darker lights, but
intensity precision is still much weaker than color recovery.

\section{GT-Driven Lighting Ablation}
\label{app:lighting_ablation}

To isolate the UE-to-Blender light adapter and the lighting stage, we use an
office scene dominated by daylight and SkyLight, with two indoor PointLights
providing local gains.  Starting from the same geometry-refined scene, we compare
four variants: no reconstructed parametric lights, GT parametric lights imported
through the adapter and refined by the lighting loop, the same light set with
energy set to zero, and a perturbation of a single light.  The variants show that
local lights have visible, independently editable effects, while all lighting
runs converge without triggering geometry rollback.

\section{VLM Prompt for Per-Object Relabeling}
\label{app:vlm_label}

Section~\ref{sec:dataset_label_light} describes isolated-object rendering plus
GPT-5.4-mini relabeling.  We list the two prompt templates used in
Lumera-2K to make the ``image evidence first, naming hints second'' policy
explicit.

\paragraph{Full prompt.}
The full prompt asks for four fields, \emph{label}, \emph{confidence},
\emph{description}, and \emph{reason}, and is used when fine-grained audit is
needed:
\begin{quote}\small
\textbf{System.} You label isolated Unreal Engine mesh asset preview images for
a 3D training dataset. The preview image is the primary evidence. Treat actor,
component, and asset names as weak hints because project naming can be noisy or
misleading. Name the visible real thing as specifically as the image supports.

\textbf{User.} Identify the visible mesh asset's visual object category. Return
only strict JSON with keys: \texttt{label}, \texttt{confidence},
\texttt{description}, and \texttt{reason}. The label must be concise, lowercase
English, snake\_case for multiple words, and should name the visual content
rather than a pack prefix, Unreal class, actor label, room name, LOD, or numeric
suffix.
\end{quote}

\paragraph{Compressed prompt.}
For million-scale labeling, we reduce token cost by using a one-line system
prompt and a shorter user prompt, requesting only \emph{label} and
\emph{confidence}.  This is the default for full Lumera-2K labeling:
\begin{quote}\small
\textbf{System.} Label isolated UE mesh preview images. Use the image first;
names are hints. Return JSON only.

\textbf{User.} Name the visible object's category. Use concise lowercase
snake\_case. Avoid UE/editor placeholders, pack names, room names, LODs, and
numbers. Return only JSON:\\
\hspace*{1em}\texttt{\{"label":"category","confidence":0.0\}.}
\end{quote}

\paragraph{Weak naming hints.}
Both prompts append a hint block containing asset name, asset path, actor label,
and, when available, rough 3D box and projection-location hints.  The hint block
is explicitly marked as weak evidence rather than ground truth.  When image and
name disagree, the model is instructed to trust the image, reducing leakage from
project naming conventions into $\mathcal{V}_{\text{fg}}$.

\section{Stage-Aware Closed-Loop Refinement: Additional Details}
\label{app:closed_loop}

Section~\ref{sec:method_ric} describes the overall stage-aware dual-agent loop.
This appendix provides details on (a) field-level stage scopes
$\mathcal{E}_\sigma$, (b) Generator and Verifier action spaces,
(c) the UE-to-Blender light adapter, (d) the full Verifier report schema and
edit-element tuples, (e) robust parsing of heterogeneous VLM tool calls, and
(f) stage prompts and stage transitions.

\subsection{Two-Stage Hard Scope \texorpdfstring{$\mathcal{E}_\sigma$}{E-sigma}}
\label{app:stage_scope}

The executor formalizes each stage as
$\mathcal{E}_\sigma = (\mathcal{F}_\sigma,\mathcal{A}_\sigma)$, where
$\mathcal{F}_\sigma$ is the frozen entity set and $\mathcal{A}_\sigma$ is the
allowed change set (Table~\ref{tab:stage_scope}).  After execution, a structured
scene diff is checked against the field-level whitelist.  Any out-of-scope
change triggers rollback and returns the violation list to the Generator,
implementing the negative feedback in Eq.~\eqref{eq:exec-sigma}.

\begin{table}[H]
  \centering\footnotesize
  \caption{Two-stage hard scopes $\mathcal{E}_\sigma$.  The frozen column lists
  entities that cannot be changed; the allowed column lists permitted changes.}
  \label{tab:stage_scope}
  \setlength{\tabcolsep}{3pt}
  \resizebox{\columnwidth}{!}{%
  \begin{tabular}{@{}lll@{}}
    \toprule
    Stage $\sigma$ & $\mathcal{F}_\sigma$ frozen & $\mathcal{A}_\sigma$ allowed\\
    \midrule
    Geometry &
      \{camera, lights, materials, mesh topology, object list\} &
      \{object $\theta_i,\mathbf{s}_i$\}; positions frozen by default \\
    Lighting &
      \{camera, mesh objects, object transforms, mesh topology, materials\} &
      \{existing-light parameters $\ell_j$, environment light, exposure\} \\
    \bottomrule
  \end{tabular}}
\end{table}

\subsection{Agent Action Spaces}
\label{app:action_space}

\paragraph{Generator action space $\mathcal{A}^G$.}
The Generator uses six semantic primitives: (i) \emph{initial scene bootstrap},
called once at $t=0$ to construct $s_0\!=\!\textsc{Bootstrap}(\Pi)$;
(ii) \emph{stage-aware high-level planning}; (iii) \emph{constrained code
execution and rendering}, the main action for geometry or lighting;
(iv) \emph{structured light update}, used only in the lighting stage and applied
directly from $E^{\text{lit}}_t$ tuples; (v) \emph{scene-state query}, which
returns object, light, material, and camera summaries of $s_t$; and
(vi) \emph{stage termination}.  Before every main action, the framework injects
the active tuple $(\sigma,\mathcal{E}_\sigma,\phi_\sigma)$ and applies the
guardrails of Eq.~\eqref{eq:exec-sigma}.

\paragraph{Verifier action space $\mathcal{A}^V$.}
The Verifier inherits VIGA's~\cite{yin2026viga} multi-view inspection tools:
camera-pose setting, object focus, visibility toggles, natural-language view
adjustments such as ``zoom out'' or ``rotate left,'' and scene queries.  All
Verifier actions are non-destructive; they do not write to the scene and only
produce visual observations $o_t^k$ for reasoning.

\subsection{UE-to-Blender Light Adapter}
\label{app:light_adapter}

When $\mathcal{L}_0$ comes from the GT Lumera-2K light record, each light
contains UE5-native fields such as lumen/candela units, a Kelvin-temperature
flag, SpotLight cone angles, and attenuation radius (Appendix~\ref{app:light_export}).
Before writing to Blender, \textsc{Bootstrap} performs a lightweight semantic
alignment: light identity and world pose are preserved; UE units are mapped to
the corresponding physical quantities in Blender Cycles; color is taken from RGB
or converted from Kelvin depending on the temperature flag; and SpotLight cone
and attenuation parameters are written to the corresponding Blender fields.  The
adapter is loss-preserving and never creates lights not present in
$\mathcal{L}_0$, keeping the lighting search space consistent with
Eq.~\eqref{eq:light-allow}.

\subsection{Full Verifier Report Schema \texorpdfstring{$r_t$}{r-t}}
\label{app:report_schema}

We formalize $r_t$ as an ordered 13-tuple of named fields:
\begin{multline}
\label{eq:report-tuple}
  r_t \,=\, \bigl\langle\,d_t,\,e_t,\,\sigma,\,\mathcal{E}_\sigma,\,V_t,\,
            f^{\text{geom}}_t,\,f^{\text{lit}}_t,\, \\
            E^{\text{obj}}_t,\,E^{\text{lit}}_t,\,
            h_t,\,\sigma^\star_t,\,\mathcal{E}^\star_t,\,\beta_t\,\bigr\rangle,
\end{multline}
where $d_t$ and $e_t$ are free-text visual differences and edit suggestions;
$\sigma$ and $\mathcal{E}_\sigma$ mirror the current scope; $V_t$ copies the
executor violation set; $f^{\text{geom}}_t$ and $f^{\text{lit}}_t$ are
stage-specific local diagnostics; $h_t$ lists required but missing carriers
(non-empty only in lighting); $\sigma^\star_t$ and $\mathcal{E}^\star_t$ propose
the next stage and scope; and $\beta_t\!\in\!\{0,1\}$ is the termination bit.
Table~\ref{tab:verifier_schema} reports field visibility in the two stages.

\begin{table}[H]
  \centering\footnotesize
  \caption{Field roles in the structured Verifier report $r_t$.  A circle marks
  a primary field; a bullet marks a field that should remain empty or only carry
  non-critical residual notes.}
  \label{tab:verifier_schema}
  \setlength{\tabcolsep}{4pt}
  \begin{tabular}{@{}lcc@{}}
    \toprule
    Field & Geometry & Lighting \\
    \midrule
    \emph{visual-difference}        & $\circ$ & $\circ$ \\
    \emph{edit-suggestion}          & $\circ$ & $\circ$ \\
    \emph{stage / edit-scope}       & $\circ$ & $\circ$ \\
    \emph{constraint-violations}    & $\circ$ & $\circ$ \\
    \emph{geometry-feedback}        & $\circ$ & $\bullet$ \\
    \emph{object-edit list}         & $\circ$ & $\bullet$ \\
    \emph{lighting-feedback}        & $\bullet$ & $\circ$ \\
    \emph{light-edit list}          & $\bullet$ & $\circ$ \\
    \emph{missing-required-hosts}   & $\bullet$ & $\circ$ \\
    \emph{fallback-stage}           & $\bullet$ & $\circ$ \\
    \emph{next-edit-scope}          & $\circ$ & $\circ$ \\
    \bottomrule
  \end{tabular}
\end{table}

\paragraph{Edit-element tuples.}
For each object $o_i$, $E^{\text{obj}}_t$ contains tuples
\begin{equation}
\label{eq:object-edit}
  \bigl(o_i,\,\Delta\theta_{z,i},\,\mathbf{s}^{\text{mul}}_i,\,
        m_i,\,\pi_i,\,\rho_i,\,c_i\bigr),
\end{equation}
where $\mathbf{s}^{\text{mul}}_i$ is an anisotropic scale multiplier;
$m_i$ is a coarse magnitude label, \textit{slight}, \textit{moderate}, or
\textit{substantial}, corresponding to $0.25\!\times$, $0.6\!\times$, and
$1.2\!\times$ of the box size; $\pi_i$ is priority; $\rho_i$ is visual evidence;
and $c_i$ is confidence.  For each light target, $E^{\text{lit}}_t$ contains
\begin{equation}
\label{eq:light-edit}
  \bigl(\zeta_j,\,\eta_j,\,\omega_j,\,v_j,\,\rho_j,\,c_j\bigr),
\end{equation}
where $\zeta_j$ is the target, either an existing light, the world probe, or
exposure; $\eta_j$ is the parameter, such as \textit{energy}, \textit{color}, or
\textit{direction}; $\omega_j$ is the update operator, \textit{set},
\textit{add}, or \textit{multiply}; and $v_j$ is the value.  These tuples are
isomorphic to the allowed side-effect sets $\mathcal{A}_\sigma$ in
Eq.~\eqref{eq:exec-sigma}, so $r_t$ can directly drive the next action
$a_{t+1}$.

\subsection{Robust Parsing of Heterogeneous VLM Tool Calls}
\label{app:json_repair}

Different VLM providers produce tool-call JSON with different failure modes,
including full-width quotes, Markdown code fences, missing outer braces, and
Python literals.  Before any tool call reaches the executor, we run a repair
layer: punctuation normalization, code-fence stripping, balanced-brace scanning,
Python-literal fallback, and remapping of raw JSON text into a legal tool call.
This layer lets GPT-5.2, GLM-4.6V, Qwen-VL, and similar VLMs share the same
closed-loop infrastructure.

\subsection{Stage Prompts and Stage Transitions}
\label{app:stage_prompts}

We maintain separate system prompts for $G$ and $V$ in the geometry and lighting
stages, $\{\Theta^{\text{geom}},\Theta^{\text{light}}\}$.  Switching stages
replaces only the prompt head in $M_t$; the rest of the conversation history is
kept.

\paragraph{$\Theta^{\text{geom}}$.}
The geometry prompt forbids adding or deleting objects, changing materials,
moving the camera, or editing lights.  It asks the model to treat
$E^{\text{obj}}_t$ as an ordered list of local repairs rather than a global
relayout instruction, freezes high-confidence anchor objects and support
parents, and freezes object positions $\mathbf{p}_i$ by default.

\paragraph{$\Theta^{\text{light}}$.}
The lighting prompt fixes the diagnostic order:
\emph{exposure $\to$ environment strength $\to$ key-light direction $\to$
shadow softness $\to$ color temperature $\to$ fill-light balance}.  It also
distinguishes re-enabling an existing disabled light from creating a new light;
the latter is always rejected.

\paragraph{Stage-specific stopping.}
The geometry stage does not stop early because of missing room-shell geometry.
The lighting stage stops immediately when the remaining residual is judged to be
non-lighting-dominated.  This prevents a common failure where the VLM keeps
editing lights to explain a missing wall or object.

\paragraph{Stage transition.}
When the geometry stage sets $\beta_t\!=\!1$ or exhausts $T_g$ rounds, the
framework switches to
$(\sigma,\mathcal{E}_\sigma,\phi_\sigma)\leftarrow
(\text{light},\mathcal{E}_{\text{light}},1)$ and replaces the prompt head with
$\Theta^{\text{light}}$.  Scene state $s$ and conversation history $M$ are
inherited.  This shared-history, switched-scope design avoids a cold start and
lets the Verifier refer to object identities already confirmed in the geometry
stage.

\paragraph{Sliding-window history.}
Following VIGA~\cite{yin2026viga}, we update
$M_{t+1}\!=\!\mathrm{Tail}_L(M_t\cup\{a_t,r_t\})$, keeping the system prompt and
the most recent $L$ action/report pairs.  Records adjacent to rolled-back steps
can be skipped automatically, keeping 30+ refinement rounds inside the VLM
context window.

\subsection{Pseudocode}
\label{app:closed_loop_alg}

\begin{algorithm}[H]
\caption{Stage-aware dual-agent refinement.}
\label{alg:ric}
\begin{algorithmic}[1]
\REQUIRE Reference image $I_{\text{ref}}$; prior tuple $\Pi$; round limits $T_g,T_l$; history window $L$
\STATE $s_0\!\leftarrow\!\textsc{Bootstrap}(\Pi)$;\;
  $M_0\!\leftarrow\!\{\Theta^{\text{geom}}\!,I_{\text{ref}}\!,s_0\}$
\FOR{$\sigma\in\{\text{geom},\text{light}\}$}
  \STATE Switch $\Theta^\sigma$ and $(\sigma,\mathcal{E}_\sigma,\phi_\sigma)$
  \FOR{$t=1$ \textbf{to} $T_\sigma$}
    \STATE $a_t\!\leftarrow\!G(M_t,I_{\text{ref}},s_{t-1};\,\sigma,\mathcal{E}_\sigma,\phi_\sigma)$
    \STATE $(s_t,V_t)\!\leftarrow\!\mathrm{exec}_\sigma(s_{t-1},a_t)$
      \COMMENT{Eq.~\ref{eq:exec-sigma}}
    \STATE $r_t\!\leftarrow\!V(I_{\text{ref}},s_t,V_t)$
    \STATE $M_{t+1}\!\leftarrow\!\mathrm{Tail}_L(M_t\cup\{a_t,r_t\})$
    \IF{$r_t$ triggers termination} \STATE \textbf{break} \ENDIF
  \ENDFOR
\ENDFOR
\RETURN $s_T$
\end{algorithmic}
\end{algorithm}

\section{Limitations and Future Work}
\label{app:limitations}

The conclusion summarizes the main limitations; we expand them here.

\paragraph{Box parsing is strongest overall, but not saturated.}
Lumera-Box reaches merged mAP$\uparrow$ $0.1141$, IoU-B$\uparrow$ $0.2472$,
and F-score$\uparrow$ $0.2762$
(Table~\ref{tab:bbox_main}), clearly outperforming DetAny3D, SpatialLM,
N3D-VLM, and WildDet3D on most metric-geometry and semantic scores.  However,
strict 3D overlap remains far from saturated for game-scale scenes.  One cause
is the heavy-tailed distribution induced by $63$M objects over $2{,}513$
projects.  Larger backbones, contrastive objectives, and explicit cross-instance
constraints are natural next steps.

\paragraph{Orientation and relation recovery remain weak points.}
The benchmark indicates that relation-oriented structure is not solved:
WildDet3D exceeds Lumera-Box by $0.1371$ on SRF$\uparrow$ and $0.3204$ on
anchor recall$\uparrow$ on the merged split.  In addition, SpatialLM-style
serialization represents yaw as a single location token and does not explicitly
handle symmetries, e.g., nearly square objects where $0$ and $\pm\pi/2$ may be
visually equivalent.
Future work should add yaw-symmetry classes, relation-aware losses, or
render-based comparison objectives.

\paragraph{Output validity needs model-level constraints.}
The sanitized protocol removes invalid labels, non-positive sizes, repeated IDs,
and ID fallback cases before scoring.  Although the new checkpoint improves
substantially over ckpt5000 (Table~\ref{tab:bbox_format}), constrained decoding
or a repair layer should still be added directly to the SFT inference path.

\paragraph{Light carriers and intensity remain incomplete.}
Two light sources that may affect appearance are not fully covered by the
current SFT target: SkyLight and near-camera strong lights whose source may be
outside the visible set.  IntrinsicHDR partially compensates for them, but this
limits accuracy.  Individual-light detection is also still weak
(F1$\uparrow$ $0.209$ at $0.5$\,m), and intensity has a log-scale
MAE$\downarrow$ of $0.431$, about a $2.7\times$ brightness error.
Ceiling-light meshes and procedural ceilings also
require better compatibility.  Extending supervision to all lights whose
influence regions intersect the view frustum, not only visible sources, and
adding explicit log-intensity losses are direct next steps.

\paragraph{Large unbounded scenes still drift geometrically.}
The outdoor split has Chamfer-L2$\downarrow$ around $17$\,m, indicating that metric
consistency in large unbounded spaces remains a front-end bottleneck.  Joint
camera and geometry calibration would improve both parsing and final assembly.

\paragraph{Engine generalization.}
Lumera-2K and the data pipeline are implemented for UE5.  Extending the
extraction and adapter layer to Unity, Godot, and other engines is necessary to
validate the representation beyond UE-style assets and lighting conventions.